\documentclass[lettersize,journal]{IEEEtran}
\usepackage{amsmath,amsfonts}
\usepackage{algorithmicx} 
\usepackage{algorithm}
\usepackage{algpseudocode}
\usepackage{array}
\usepackage[caption=false,font=normalsize,labelfont=sf,textfont=sf]{subfig}
\usepackage{textcomp}
\usepackage{stfloats}
\usepackage{url}
\usepackage{verbatim}
\usepackage{graphicx}
\usepackage{cite}
\usepackage{epsfig}
\usepackage{xcolor}
\usepackage{colortbl}
\usepackage{multirow}
\usepackage{multicol}
\usepackage{caption}
\usepackage[colorlinks=true]{hyperref}

\def\BibTeX{{\rm B\kern-.05em{\sc i\kern-.025em b}\kern-.08em
    T\kern-.1667em\lower.7ex\hbox{E}\kern-.125emX}}

\begin{document}

\def\x{{\mathbf x}}
\def\L{{\cal L}}

\renewcommand{\algorithmicrequire}{\textbf{Input:}}  
\renewcommand{\algorithmicensure}{\textbf{Output:}} 

\newtheorem{theorem}{Theorem}
\newtheorem{lemma}{Lemma}
\newtheorem{defn}{Definition}
\newtheorem{remark}{Remark}

\newcommand{\mb}[1]{\mathbb{#1}}
\newcommand{\mf}[1]{\mathbf{#1}}
\newcommand{\mc}[1]{\mathcal{#1}}
\newcommand{\mr}[1]{\mathrm{#1}}
\newcommand{\mrb}[1]{\mathrm{\mathbf{#1}}}

\def\eg{\emph{e.g.}}
\def\etal{\emph{et al. }}
\def\ie{\emph{i.e.}}

\definecolor{mygray}{gray}{.9}
\definecolor{myyellow}{RGB}{251,255,216}

\def\red#1{\textcolor{red}{#1}}
\def\blue#1{\textcolor{blue}{#1}}
\def\green#1{\textcolor{green}{#1}}

\title{ForensicsForest Family: A Series of Multi-scale Hierarchical Cascade Forests for Detecting GAN-generated Faces}

\author{Jiucui Lu, Jiaran Zhou, Junyu Dong, Bin Li, Siwei Lyu, Yuezun Li
\thanks{{\em Corresponding authors}: Jiaran Zhou and Yuezun Li.}
\thanks{Jiucui Lu, Jiaran Zhou, Junyu Dong, and Yuezun Li are with the College of Computer Science and Technology, Ocean University of China, Qingdao, China. e-mail: lujiucui@stu.ouc.edu.cn;\{zhoujiaran;dongjunyu;liyuezun\}@ouc.edu.cn. }
\thanks{Bin Li is with Guangdong Key Laboratory of Intelligent Information Processing, Shenzhen Key Laboratory of Media Security, Shenzhen University, Shenzhen 518060, China. Email: libin@szu.edu.cn}
\thanks{Siwei Lyu is with the University at Buffalo, SUNY, USA. Email: siweilyu@bufflao.edu.}}

\markboth{Journal of \LaTeX\ Class Files,~Vol.~14, No.~8, August~2021}%
{Shell \MakeLowercase{\textit{et al.}}: A Sample Article Using IEEEtran.cls for IEEE Journals}


\maketitle

\begin{abstract}
The prominent progress in generative models has significantly improved the authenticity of generated faces, raising serious concerns in society. To combat GAN-generated faces, many countermeasures based on Convolutional Neural Networks (CNNs) have been spawned due to their strong learning capabilities. In this paper, we rethink this problem and explore a new approach based on forest models instead of CNNs. Concretely, we describe a simple and effective forest-based method set, termed {\em ForensicsForest Family}, to detect GAN-generate faces. The ForensicsForest family is composed of three variants: {\em ForensicsForest}, {\em Hybrid ForensicsForest} and {\em Divide-and-Conquer ForensicsForest}. ForenscisForest is a novel Multi-scale Hierarchical Cascade Forest that takes appearance, frequency, and biological features as input, hierarchically cascades different levels of features for authenticity prediction, and employs a multi-scale ensemble scheme to consider different levels of information comprehensively for further performance improvement. Building upon ForensicsForest, we create Hybrid ForensicsForest, an extended version that integrates the CNN layers into models, to further enhance the efficacy of augmented features. Furthermore, to reduce memory usage during training, we introduce Divide-and-Conquer ForensicsForest, which can construct a forest model using only a portion of training samplings. In the training stage, we train several candidate forest models using the subsets of training samples. Then, a ForensicsForest is assembled by selecting suitable components from these candidate forest models.
Our method is validated on state-of-the-art GAN-generated face datasets and compared with several CNN models, demonstrating the surprising effectiveness of our method in detecting GAN-generated faces. 
\end{abstract}

\begin{IEEEkeywords}
Digital forensics, GAN-generated face detection, Random forest
\end{IEEEkeywords}
\section{Introduction}

\begin{figure}[!t]
\centering
\includegraphics[width=\linewidth]{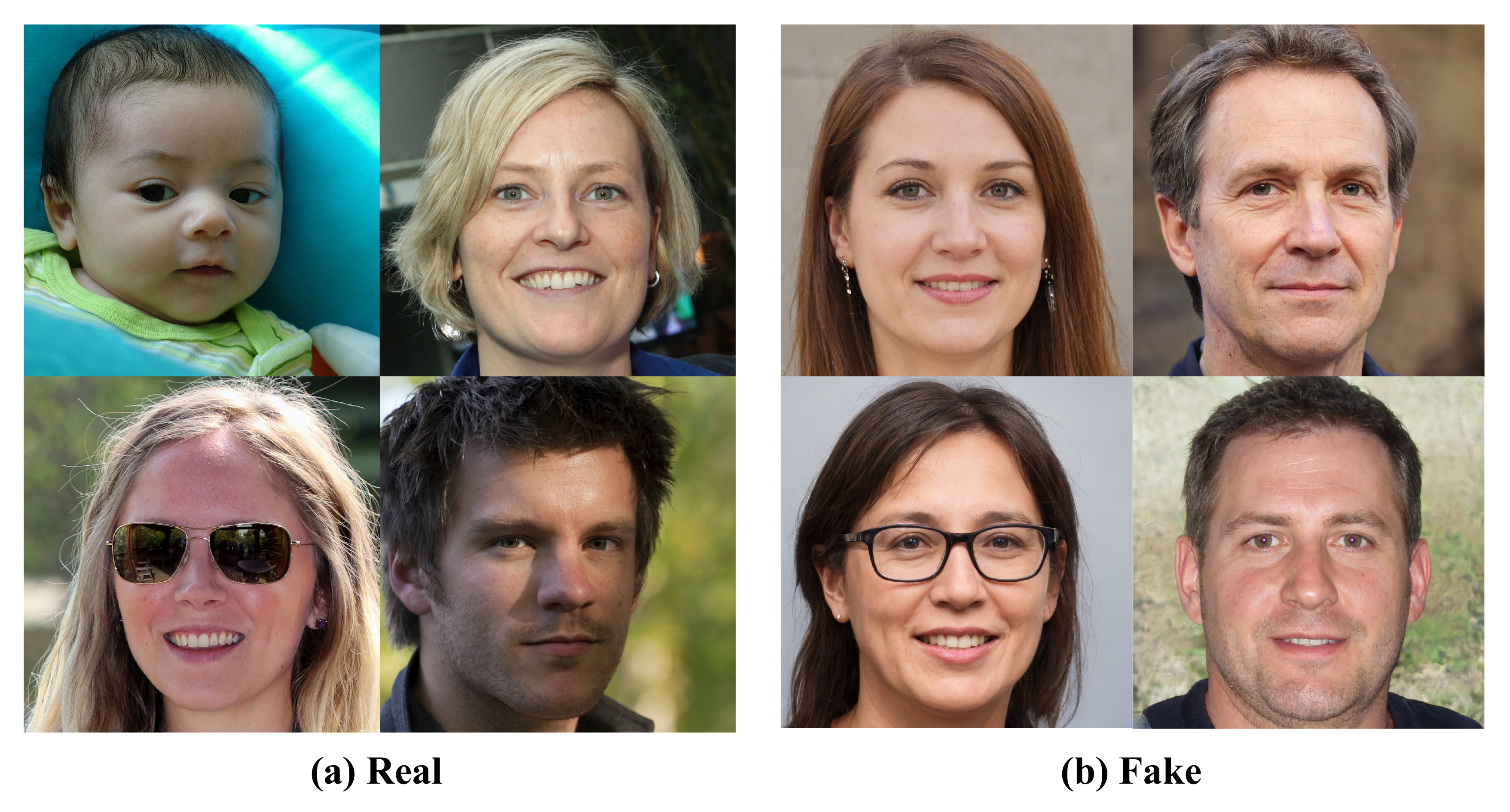}
\vspace{-0.7cm}
\caption{\small Examples of real and GAN-generated faces (selected from StyleGAN2 \cite{karras2020analyzing}).}
\label{fig:ganexamples}
\vspace{-0.6cm}
\end{figure}

\IEEEPARstart{F}{ace} forgery has significantly advanced in quality and efficiency, thanks to the advent of deep generative models (\eg, GAN \cite{karras2017progressive, goodfellow2020generative, karras2020analyzing}, VAE \cite{kingma2013auto}). As shown in Fig.~\ref{fig:ganexamples}, the GAN-generated faces exhibit a high level of realism, which can hardly be distinguished by human eyes. Since human faces are important biometrics, the abuse of GANs can raise a severe security concern for society, \eg, forging a fake identity on social platforms, deceiving users for fraud, etc \cite{farid2022creating}. As such, detecting the GAN-generated faces is of great importance.

The current GAN-generated faces detection methods \cite{wang2020cnn, liu2020global, guo2022robust} are mainly based on Convolutional Neural Network (CNN) models for their powerful learning abilities demonstrated in various vision tasks. With the availability of large-scale datasets of face forgeries \cite{rossler2019faceforensics++,li2020celeb}, it is possible to design new complex forms of CNN architectures with more parameters, without the risk of overfitting. However, despite these CNN-based methods having shown promising performance, they have two significant limitations that may obstruct their application in daily practice: 1) High demand for computing resources. Since the CNN models usually contain plenty of weight parameters, training them requires careful fine-tuning and expensive computing resources, \eg, GPUs; 2) Security concerns. It has been proven that CNN-based methods are vulnerable to adversarial attacks, which can mislead the prediction by only adding imperceptible noises to input faces \cite{carlini2020evading,gandhi2020adversarial}. It works because a large number of weight parameters makes the classification boundary more complicated, increasing the possibility of pushing a sample crossing the border with less effort. Also, due to the differentiability of CNNs, the attacks can be easily achieved by optimizing an objective function. These limitations drive us to think can the models being low resource demand and non-differentiable be used to detect GAN-generated faces?

\begin{figure*}[t]
\centering
\includegraphics[width=\linewidth]{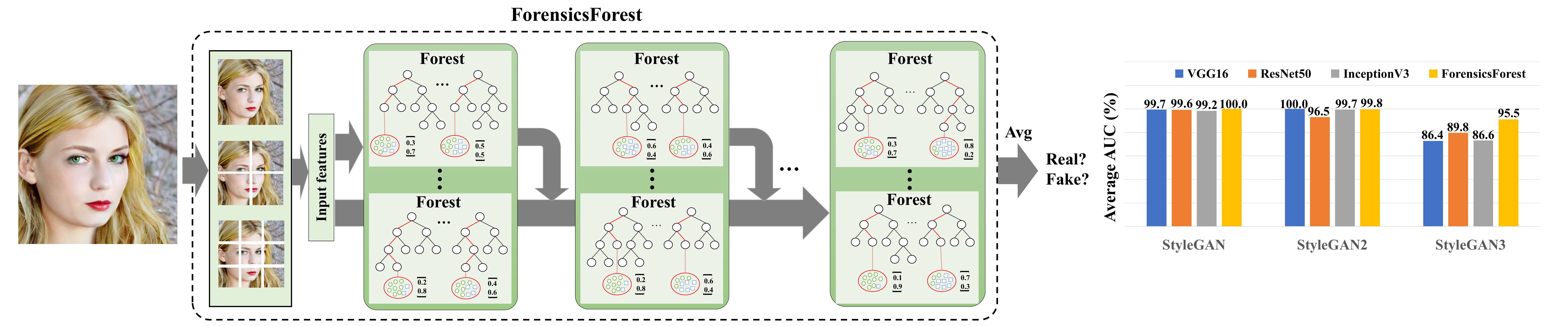}
\vspace{-0.6cm}
\caption{\small \small Diagram of the proposed ForensicsForest, the core of {\em ForensicsForest Family} for detecting GAN-generated faces. Our method can achieve competitive and even better performance compared to CNN-based detection methods. Other two variants, Hybrid ForensicsForest and Divide-and-Conquer ForensicsForest are developed based on ForensicsForest. The details of each variant can be seen in Section \ref{sub:ForensicsForest}, Section \ref{sub:Hybrid-ForensicsForest}, and Section \ref{sub:Divide-and-Conquer} respectively.}
\vspace{-0.3cm}
\label{fig:framework}
\end{figure*}

\begin{figure}[t]
\centering
\includegraphics[width=\linewidth]{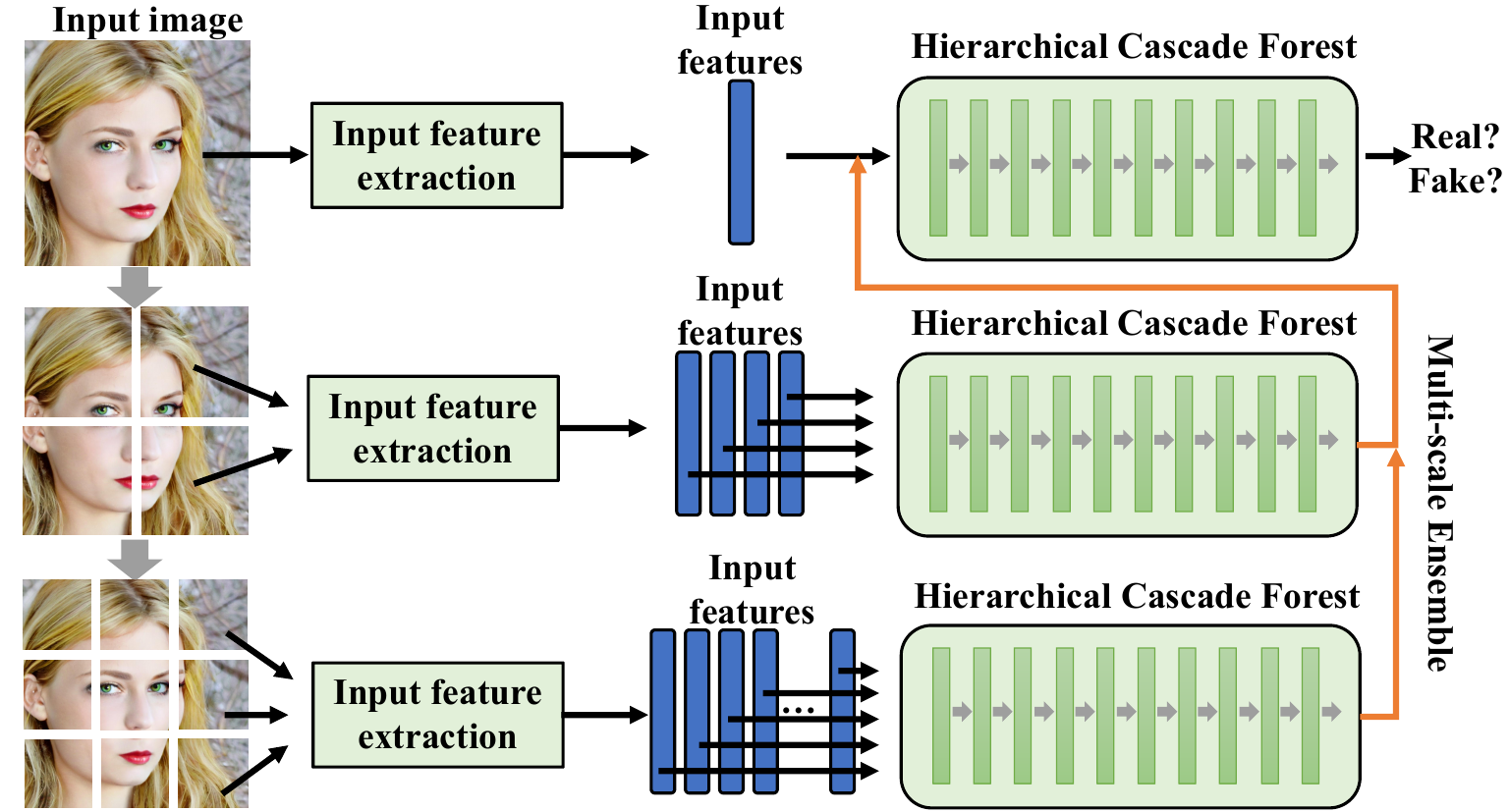}
\vspace{-0.6cm}
\caption{\small Overview of the proposed multi-scale hierarchical cascade forest.}
\label{fig:overview}
\vspace{-0.3cm}
\end{figure}

In this paper, we rethink this problem and describe a simple and effective method set called {\em ForensicsForest Family} to expose GAN-generated faces (see Fig.~\ref{fig:framework}). We adopt the forest model as the base in replace of CNN models to overcome the aforementioned limitations, as the forest is decision-based, which contains few weight parameters, and is not differentiable, thus naturally resisting a set of general adversarial attacks \cite{carlini2020evading,gandhi2020adversarial}. The proposed ForensicsForest family is composed of three variants, which are {\em ForensicsForest}, {\em Hybrid ForensicsForest} and {\em Divide-and-Conquer ForensicsForest} respectively.
ForensicsForest is a newly proposed architecture called Multi-scale Hierarchical Cascade Forest, containing three main components: Input Feature Extraction, Hierarchical Cascade Forest, and Multi-scale Ensemble, respectively (see Fig.~\ref{fig:overview}). The input feature extraction is a preprocessing step to extract informative features with fixed dimensions, enabling our detector to be independent of image size. Concretely, we first split the input image into multiple patches and then extract two types of features for each patch: the color histogram as appearance features and the power spectrum as frequency features. To capture the characteristics of faces, we additionally extract the facial landmarks as biological features. This is because the topology of the facial landmarks is typically not taken into account in the generation of faces using GANs, which can result in inconsistency with the facial landmarks found in real faces~\cite{yang2019exposing_landmark}. These features are concatenated and sent into a hierarchical cascade forest for prediction. The hierarchical cascade forest is composed of several cascade forest layers, where the features of each patch are hierarchically integrated into consecutive layers. This design allows for alternatively processing each patch, augmenting the features of each patch by incorporating the knowledge of the previous patch while minimizing the computational overhead. Moreover, we propose a multi-scale ensemble approach that takes features of different scales by adjusting the sizes of patches. This enables the model to effectively learn the discriminative features and combines the obtained results for final prediction. 

Expanding on the concept of ForensicsForest, we propose Hybrid ForensicsForest as an extension that explores the feasibility of integrating CNN layers into the forest model. This is motivated by the proven effectiveness of CNN in learning tasks. By combining CNN with the forest model, we aim to enhance the output of forest layers and improve the effectiveness of augmented features. In addition, constructing forest-based models typically requires loading all training samples into memory simultaneously, leading to significant memory costs. As such, we further propose the Divide-and-Conquer ForensicsForest to address the issue of high memory cost while maintaining favorable detection performance. In this approach, we divide the training stage into two steps. Firstly, we train several candidate forest models using only a small portion of training samples. This allows us to reduce the memory requirement during training. Next, we develop a strategy to select the best components from these candidate forest models and assemble these components as a final forest model. This final model retains the favorable detection performance while significantly reducing the resource costs associated with using all training samples at once.

We conduct extensive experiments on various types of state-of-the-art GAN-generated faces, and compare our method with several counterparts, showing that our method is surprisingly effective in exposing GAN-generated faces in comparison to CNN-based methods. Moreover, we thoroughly study the effect of each component of our method, as well as the robustness and generalization ability.

The contribution of this paper is summarized as follows
\begin{enumerate}
    \item We describe a new forest-based method named {\em ForensicsForest}, which is a novel Multi-scale Hierarchical Cascade Forest to process face patches hierarchically in a multi-scale ensemble manner. To the best of our knowledge, we are the first to investigate the potential of forest models to expose GAN-generated faces, offering fresh insights for future research. 

    \item Building upon ForensicsForest, we develop {\em Hybrid ForensicsForest}, which explores the integration of CNN layers into forest layers, aiming to further enhance the effectiveness of augmented features.

    \item To reduce the memory cost, we describe {\em Divide-and-Conquer ForensicsForest}, an approach that selects the most suitable components from a set of candidate forest models which are trained using a small portion of samples, and assemble these components as a final ForensicsForest. 

    \item We extensively evaluate our method on recent GAN-generated faces, including StyleGAN \cite{Karras_2019_CVPR}, StyleGAN2 \cite{karras2020analyzing}, and StyleGAN3 \cite{karras2021alias}. We compare our method with CNN-based counterparts and thoroughly analyze the effect of various modules in our method. Furthermore, we validate the performance of our method on robustness, cross-datasets, and other generative models, such as ProGAN \cite{karras2018progressive}, StarGAN \cite{choi2018stargan} and Diffusion model \cite{rombach2022high}.
\end{enumerate}

This paper is extended from our {\tt ICME conference paper \cite{lu2023fforest} (Oral)} in several aspects: (1) We improve the input feature extraction module by adding facial landmarks as biological features; (2) We propose Hybrid ForensicsForest, which integrates the CNN layers into ForensicsForest, and validate its performance on GAN-generated face datasets; (3) We then develop Divide-and-Conquer ForensicsForest, which selects the suitable components from a set of candidate forest models to form a final model, and investigate its performance with various settings; (4) We conduct more experiments on the comparison with dedicated GAN-generated face detection methods, and study the ability of our method on different datasets and different generative models.


\section{Background and Related Works}
\label{sub:background}

\smallskip
\noindent{\bf GAN-generated Faces Detection.} In the early stage, many forensic methods are based on statistic signals to expose forged images \cite{prasad2006resampling,kirchner2008fast,mahdian2008blind}. However, with the advent of GAN models, the faces can be synthesized in an end-to-end fashion with very high visual quality, and these early methods are no longer effective in detecting GAN-generated faces. 

Recently, CNNs have been widely employed in detection methods due to their excellent performance on vision tasks, in order to expose subtle forgery traces. Some methods learn the detectable clues by directly training CNN models with vanilla or self-designed architectures in a supervised way, \eg, \cite{boulkenafet2016face,wang2020cnn,hulzebosch2020detecting,patchforensics,liu2020global,bai2021robust,wang2021representative,zhao2021multi,han2023fcd}. There are also many methods relying on empirically selected clues in the following aspects: 1) Physiological signals. Li \etal \cite{li2018ictu} propose a Long-term Recurrent Neural Network (LRCN) to detect the inconsistency in eye blinking. Yang \etal \cite{yang2019exposing} expose the head pose inconsistency to expose generated faces. Guo \etal \cite{guo2022robust} find that the generated faces do not consider the consistency in eye corneal specular highlight and use it as clues to expose generated faces. Similarly, other physiological signals such as remote photoplethysmography (rPPG) are also used in \cite{chen2022pulseedit} to exhibit the inconsistency between real and fake. 2) Artifact signals. Several works find that the upsampling process in GAN face generation can leave specific artifacts \cite{zhang2019detecting,durall2020watch}, and the blending artifacts are found in face swapping operations \cite{li2019exposing,li2020face}. More recently, LGrad \cite{tan2023learning} utilize gradient artifacts as the general clues for detection. 3) Spatial or frequency signals. Several works detect the inconsistency in different color domains, \eg, \cite{mccloskey2018detecting,matern2019exploiting,li2020identification}, while other works attempt to extract traces in the frequency domain, \eg, \cite{durall2019unmasking,qian2020thinking,li2021frequencyaware,liu2021spatialphase}. 
Extracting these clues relies on CNNs or manually designed modules, and then sending these clues into CNNs for final prediction. In contrast to these recent methods, we investigate the feasibility of using a non-CNN model and describe a new series of forest-based methods to expose GAN-generated faces.

\smallskip
\noindent{\bf Deep Forest.}
The forest is the classical decision model for classification. Due to the nature of decision models, they are not differentiable. In contrast, Deep Neural Networks (DNN) models are differentiable networks with deeper architectures and multiple layers of differentiable parameterized modules. Inspired by the success of DNN models, Deep Forest \cite{zhou2017deep} is proposed in a non-DNN style deep model with a cascade structure. Specifically, Deep Forest is a layer-by-layer architecture and each layer contains multiple decision trees. The output of each layer is concatenated with the input and sent into the next layer. Due to this novel design, it performs better than the traditional decision models (\eg, Random Forests, XGBoost, etc), and achieves competitive, sometimes better performance than CNN models. Compared to DNNs, in addition to the non-differentiable property, Deep Forest requires much fewer hyper-parameters, and its model complexity can be automatically determined in a data-dependent way. However, it usually focuses on the general classification in small scales (\eg, CIFAR10 \cite{krizhevsky2009learning}, MINIST \cite{lecun1998gradient}), which can hardly be directly applied to the task of GAN-generated face detection. The difference between our method and Deep Forest is elaborated in Section~\ref{sub:comparsion}.

\section{ForensicsForest}
\label{sub:ForensicsForest}

\subsection{Model Architecture}
Our method contains three components: Input Feature Extraction, Hierarchical Cascade Forest, and Multi-scale Ensemble. We introduce each component in a sequel. 



\begin{figure}[t]
\centering
\includegraphics[width=\linewidth]{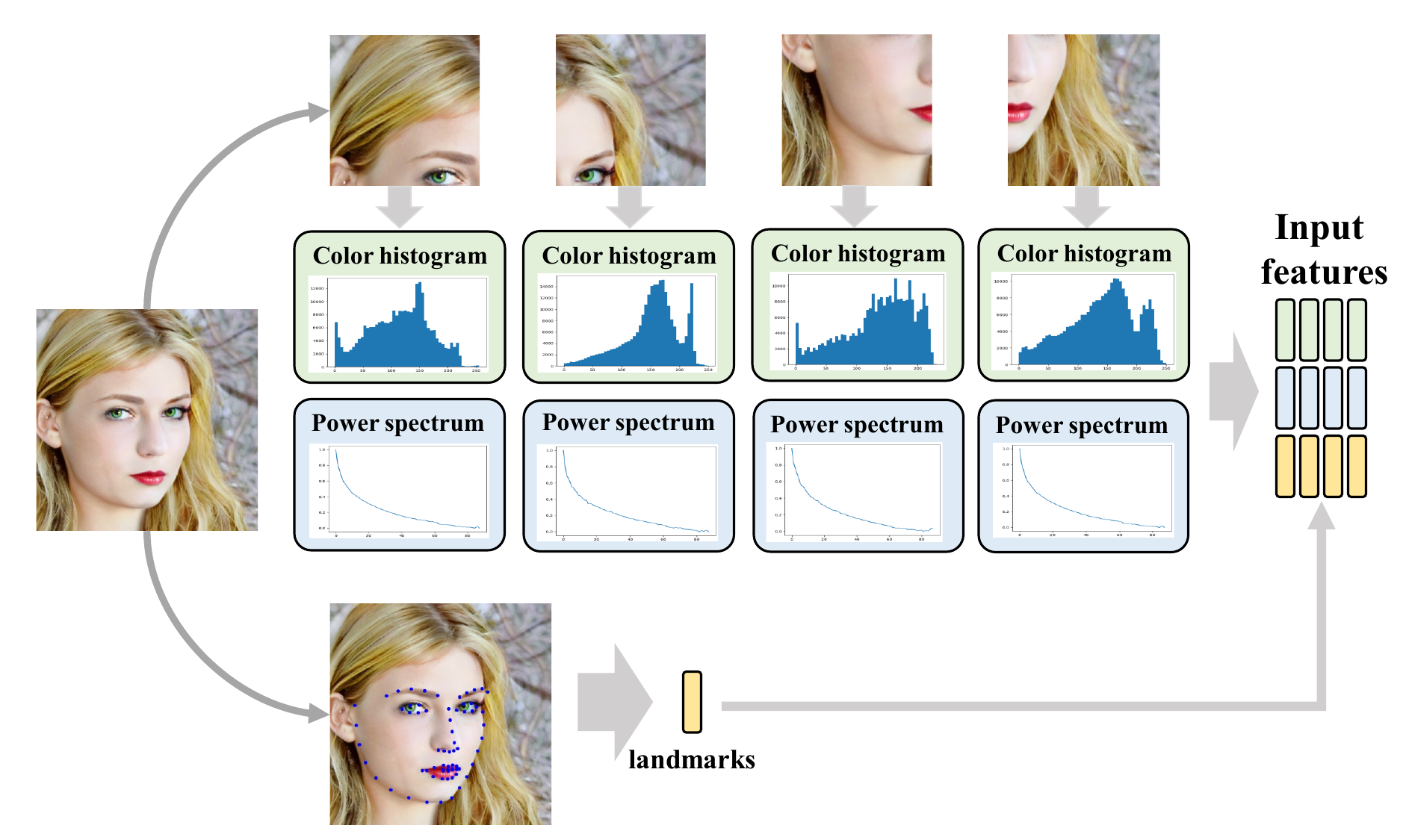}
\vspace{-0.7cm}
\caption{\small Illustration of input feature extraction. We extract the color histogram, power spectrum, and facial landmarks as the appearance, frequency, and biology features respectively. See text for details. }
\label{fig:infeat}
\vspace{-0.3cm}
\end{figure}

\smallskip
\noindent{\bf Input Feature Extraction.} Instead of using the whole image as input, we use the features extracted from the image. Note that the extracted features are in fixed dimensions, regardless of the input image size. Thus our method is feasible to handle the input images of arbitrary size. Specifically, we extract three types of features: appearance features, frequency features, and biology features.

For appearance features, we employ simple color histograms. Concretely, we first make a color histogram for each channel and flatten them together as the appearance features (3 × 256). Moreover, we extract frequency features as complementary to appearance features. We first convert each channel of the input image into a frequency map using the Fast Fourier Transform (FFT) and then transform the frequency map into a power spectrum using the azimuthal average as the frequency features (3 × 88). For biology features, we extract the facial landmarks and use the coordinates of these landmarks to represent the characteristics of faces (2 × 68). Fig.~\ref{fig:infeat} shows the overview of input feature extraction.

In a general formulation, we can extract appearance and frequency features from $N (N \geq 1)$ patches of input images, where $N = 1$ indicates the two features are extracted from the whole image, and extract biology features from the whole image. 

\begin{figure*}[!th]
\centering
\includegraphics[width=0.95\linewidth]{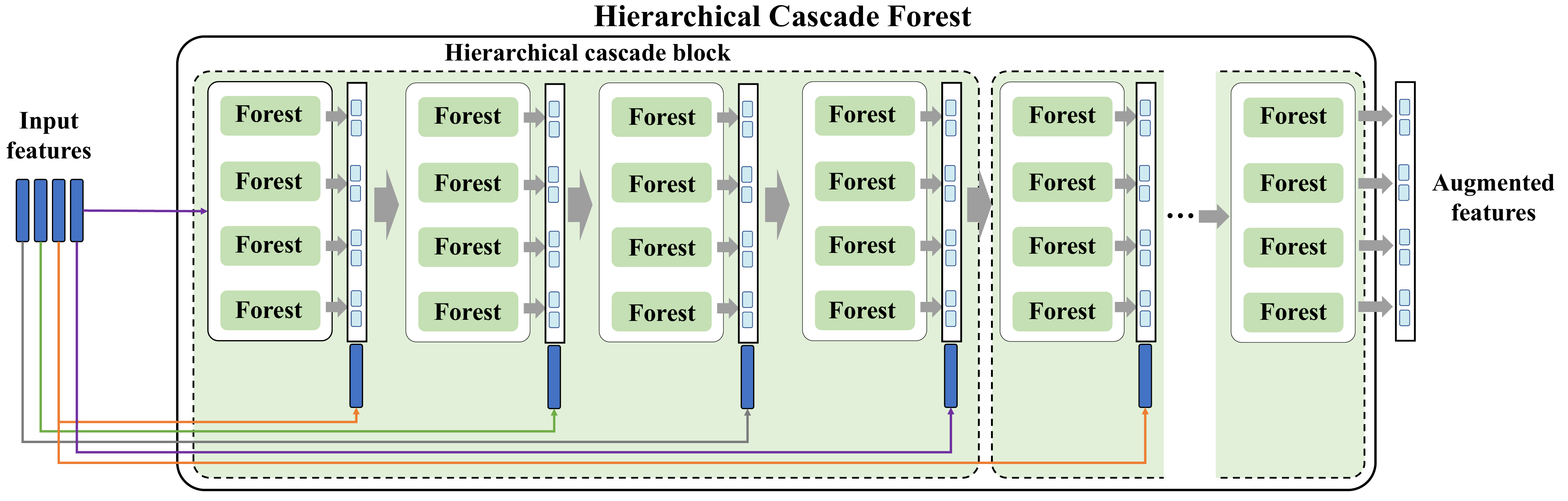}
\vspace{-0.3cm}
\caption{\small Overview of the hierarchical cascade forest. The input features are hierarchically sent into the corresponding layer in a hierarchical cascade block.  See text for details.}
\label{fig:hcforest}
\vspace{-0.3cm}
\end{figure*}

\smallskip
\noindent{\bf Hierarchical Cascade Forest.}
Given the extracted input feature vectors, we develop a hierarchical cascade forest to determine the authenticity. As shown in Fig.~\ref{fig:hcforest}, this forest contains multiple hierarchical cascade blocks, where each block is composed of $N$ cascade forest layers. Each layer contains two random forests and two completely random forests. The random forest is composed of CART decision trees which are created using the Gini index to select the best feature attributes for a partition. By contrast, the completely random forest consists of completely random trees created by randomly selecting feature attributes. In our task, we have two classes to predict, \ie, real or fake. Thus each forest generates a two-dimension probability vector. The concatenation of these vectors inside a layer is the augmented feature. Specifically, for each random forest, we use {$100$} random trees and the same setting for each completely random forest, which contains {$100$} completely random trees. The workflow is as follows: the output of $i$-th layer is the augmented feature, which is then concatenated with the feature vector of $(i+1)$-th patch as the input of $(i+1)$-th layer. For the first block, the input of the $1$-st layer is the input feature of the $1$-st patch. {For other blocks, the input of the $1$-st layer is the concatenation of the input feature of the $1$-st patch and the augmented feature from the last block.}

\smallskip
\noindent{\bf Multi-scale Ensemble.} 
The output of the hierarchical cascade forest can be used for the final prediction. However, only relying on one hierarchical cascade forest overlooks the forgery information in other scales. Hence, we propose a multi-scale ensemble, which incorporates the forgery information of multiple scales as the final feature for prediction. As shown in Fig.~\ref{fig:overview}, we construct several hierarchical cascade forests and each forest corresponds to one patch number (\eg, $N=1,4,9$). Specifically, we use the forest for the whole image ($N=1$) as the base forest. {The input features from this forest are concatenated with the augmented features from other forests ($N>1$) as the final input.} By considering multiple scales, the detection performance is further improved.

\subsection{Training and Inference}
\label{sub:training}
\noindent{\bf Training.} Different from CNN models, our method is constructed on decision trees. Thus the training process is not to learn parameters, but to create the forest structure directly on given training samples. Note that our method contains two types of forests, random forest and completely random forest. Denote the training set as $\mc{D}$, and $(x, y) \in \mc{D}$ as a training sample. Note that $y \in \{0, 1\}$, where $0$ denotes real and $1$ denotes fake. The Gini index of $\mc{D}$ is defined as 
\begin{equation}
    \text{Gini}(\mc{D}) = 2p(1 - p),
\end{equation}
where $p$ is the probability of samples being labeled $1$. 
Let $\mc{A}_i$ be an attribute from feature $\mc{A}$ that has possible values in set $\mc{V}$. Assume $a \in \mc{V}$ is one of the possible values. The Gini index of $\mc{A}_i = a$ can be defined as
\begin{equation}
    \begin{array}{c}
         \text{Gini}(\mc{D}, \mc{A}_i = a) = \frac{|\mc{D}_1|}{|\mc{D}|} \text{Gini} (\mc{D}_1) + \frac{|\mc{D}_2|}{|\mc{D}|} \text{Gini} (\mc{D}_2), \\ \vspace{0.1cm}
          \mc{D}_1 = \{(x, y) \in \mc{D} | \mc{A}_{i}(x) = a \}, \mc{D}_2 = \mc{D} - \mc{D}_1,
    \end{array}
\end{equation}
The creation of decision trees is to find the best partition recursively, as 
\begin{equation}
    \mathop{\arg\min}\limits_{a\in \mc{V}, \mc{A}_i \in \mc{A}} \text{Gini}(\mc{D}, \mc{A}_i = a).
\end{equation}
These forests randomly select $\sqrt{d}$ candidate feature attributes, where $d$ is the number of feature attributes. To mitigate the risk of overfitting, each forest is created using $k$-fold cross-validation, that is first to randomly divides $\mc{D}$ into $k$ sets without overlap, and then use $k-1$ sets for training and rest set for validation. This process is repeated $k$ times until each set has been used for validation. 

In contrast to CNN models, we dynamically decide the number of cascade layers in ForensicsForest. The criterion for terminating the training process is as follows: 1) Meeting the maximum number of layers. In order to control resource consumption, the forest stops growing if the number of layers is greater than the maximum, or 2) The performance is not improved anymore even if more layers are added. Specifically, we constantly evaluate the output of each layer by feeding training data. If a predefined number of layers behind the current layer can not further improve the performance, we do not add these layers into the forest and terminate the training process. 

\smallskip
\noindent{\bf Inference.} Once the forest is constructed, we can send testing face images to it and average the output of each forest of the last layer to obtain the final prediction.

\subsection{Comparing with Classic Deep Forest}
\label{sub:comparsion}
\noindent{\bf Input Feature Extraction.}
The classic Deep Forest is designed for low-dimensional images, \eg, MINIST ($28 \times 28$), CIFAR ($64 \times 64$) dataset, which can not handle the high-dimensional images such as GAN-generated face images, \eg, StyleGAN ($1024 \times 1024$), due to the significant high resource consumption. Thus, we convert the GAN-generated face images into low-dimensional features as the input. The advantages are that the resource consumption is independent of the dimension of images, and redundant content can be discarded.     

\begin{figure*}[t]
\centering
\includegraphics[width=\linewidth]{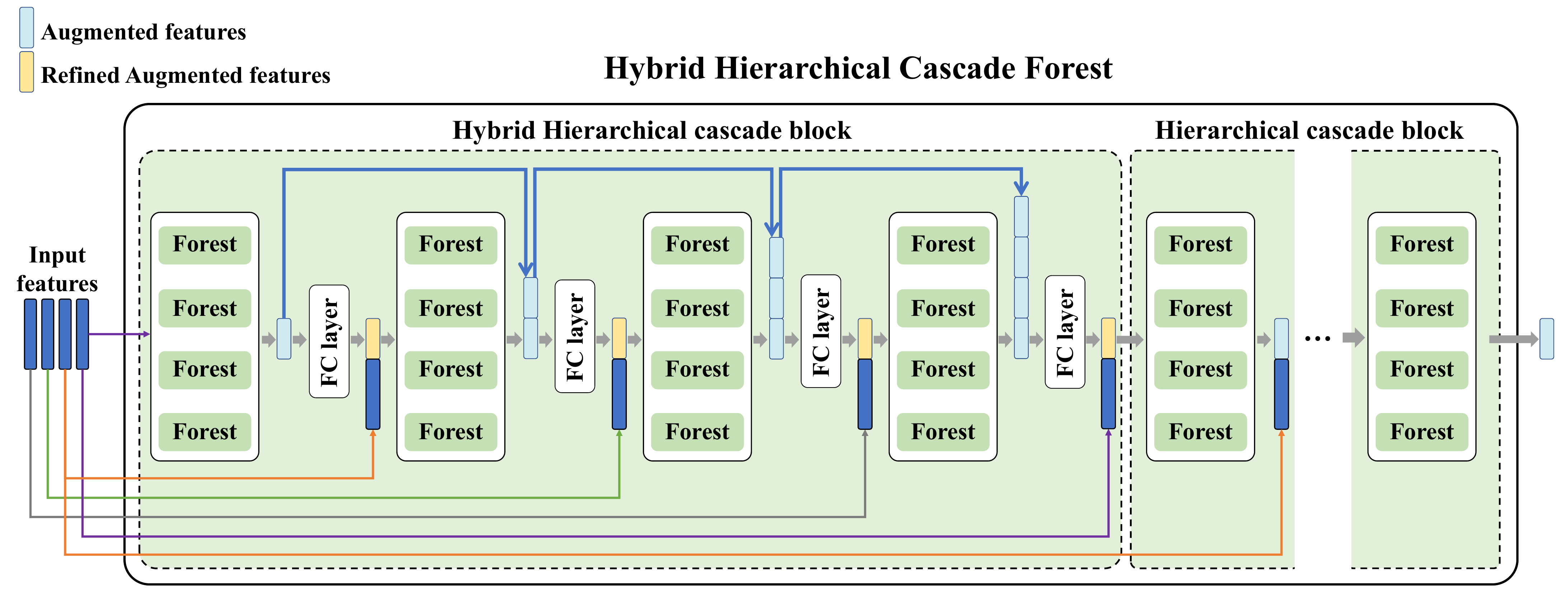}
\vspace{-0.6cm}
\caption{\small Overview of the proposed Hybrid ForensicsForest. In contrast to ForensicsForest, Hybrid ForensicsForest integrates FC layers into the model, appending a FC layer behind the augmented features from the forest layer and refining the features for the next layer.}
\label{fig:hybrid}
\vspace{-0.2cm}
\end{figure*}

\smallskip
\noindent{\bf Hierarchical Cascade Forest.} 
The classic Deep Forest uniformly cascades the input features with the augmented features out from each layer. Thus the computation cost is positively correlated with the size of features. In contrast, we propose a hierarchical cascade, which first splits the input features into different pieces (\eg, $4$ pieces in Fig.~\ref{fig:hcforest}), and alternatively cascades different pieces with the augmented features out from each layer. In this way, the computation cost is greatly reduced, and more importantly, the augmented features of one layer can absorb the knowledge of the previous piece of feature, which can learn local associations between pieces while obtaining the global feature ultimately.

\smallskip
\noindent{\bf Multi-scale Ensemble.} The classic Deep Forest does not consider the multi-scale process. However, since the GAN-generated face images are usually in high dimensions containing more complex content, only using one scale is ineffective. Thus we propose a multi-scale ensemble, which considers different scales of information, by fusing the augmented features from different scales together for the final prediction.

\section{Hybrid ForensicsForest}
\label{sub:Hybrid-ForensicsForest}

The effectiveness of CNNs in various tasks has been well-established, mainly due to their learnable parameters. Inspired by this, we aim to investigate the potential of integrating these learnable parameters into the forest model. This integration seeks to strike a balance between CNNs and decision-based models. Building on the proposed ForensicsForest, we describe Hybrid ForensicsForest, which integrates the CNN layers into the forest model layers, to further refine the augmented features in ForensicsForest.

In our approach, we employ the Fully Connected (FC) layer to enhance the forest features. As shown in Fig.~\ref{fig:hybrid},
we append a FC layer after each forest layer and send the refined augmented features into the next layer. We opt for the FC layer instead of the Convolution layer because {the augmented features derived from the forest are} compact with low dimensions, making them more amenable to global refinement. To maximize the potential of the FC layer, we propose an iterative cascade strategy, which stacks the augmented features of the previous layer to the current layer (Blue arrows in Fig.~\ref{fig:hybrid}). This allows for the collection of more information to be used for the FC layer. More strategies involving the use of FC layers can be found in Section~\ref{sub:hybridff}.

\begin{figure}[t]
\centering
\includegraphics[width=0.8\linewidth]{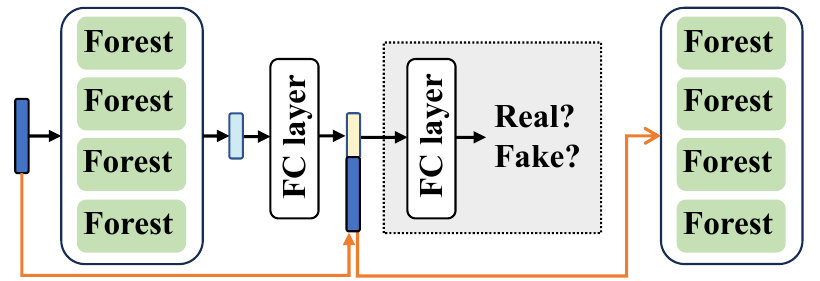}
\vspace{-0.2cm}
\caption{\small Illustration of the training process of Hybrid ForensicsForest. In training, we add an extra FC layer (shown in the grey box) to implicitly guide the learning. In testing, we only use the first FC layer.}
\label{fig:hybrid-training}
\vspace{-0.2cm}
\end{figure}

Training this Hybrid ForensicsForest is non-trivial, as it involves combining a decision model with learnable parameters. This makes it challenging to optimize the model in an end-to-end fashion using the existing training schemes for ForensicsForest. 
To address this challenge, we develop a hybrid training scheme that can jointly construct the forests and learn the parameters of FC layers. Specifically, we first construct the forest layer using the training scheme described in Section~\ref{sub:training}. Then we train the appended FC layers. However, training the FC layer within the forest model poses difficulties as the gradients fail to propagate through the model. Consequently, it is not feasible to optimize the FC layer using a task-related objective function (\eg, binary classification) based on the input and output of the model. 
To overcome this limitation, we train the FC layer in each forest layer separately. We utilize the augmented features from the forest layer as training samples for the FC layer and introduce a second FC layer as an auxiliary component to guide the learning of the first FC layer. The objective for training these two FC layers is a simple binary cross-entropy loss, \ie, distinguishing between real and fake. In this way, the first FC layer learns to align the augmented features towards a distinguishable distribution. Once training is complete, we retain only the first FC layer and disable the second FC layer during the testing phase. The training procedure is depicted in Fig.~\ref{fig:hybrid-training}.

Note that we do not directly add the FC layers on images. Instead, we insert them in an alternative manner among the forest layers. Despite the FC layers being differentiable, the gradients cannot be back-propagated to the input image due to the inherent nature of forests. Thus the proposed Hybrid ForensicsForest remains resilient against general adversarial attacks.


\section{Divide-and-Conquer ForensicsForest}
\label{sub:Divide-and-Conquer}

Unlike the batch-style training scheme commonly used in CNN, ForensicsForest requires loading all training samples into memory at once to construct the forest structure. However,
this training scheme poses limitations when applied in the wild, as the ForensicsForest model may fail to be constructed if the available memory size is insufficient to accommodate all training samples. To address this issue, one possible solution is to only use a portion of the training samples. However, this approach inherently lacks the knowledge from the excluded training samples. To overcome this issue, we propose Divide-and-Conquer ForensicsForest which can greatly reduce the memory cost while retaining the favorable detection performance.

In the training stage, we randomly divide the training samples into two portions and use one portion to train a ForensicsForest. By utilizing only a part of the training samples, the memory required for constructing the forest is significantly reduced. This training process is repeated several times, resulting in a collection of forest models, known as candidate forest models. Since each forest model is trained using a randomly divided portion of training samples, it can capture a specific perspective knowledge of all training samples. We believe that by carefully designing a strategy, we can integrate these different perspectives of knowledge from the forest models into a final one. This integration aims to be equivalent to using all training samples at once. This strategy is elaborated as follows.

\begin{figure}[t]
\centering
\includegraphics[width=0.95\linewidth]{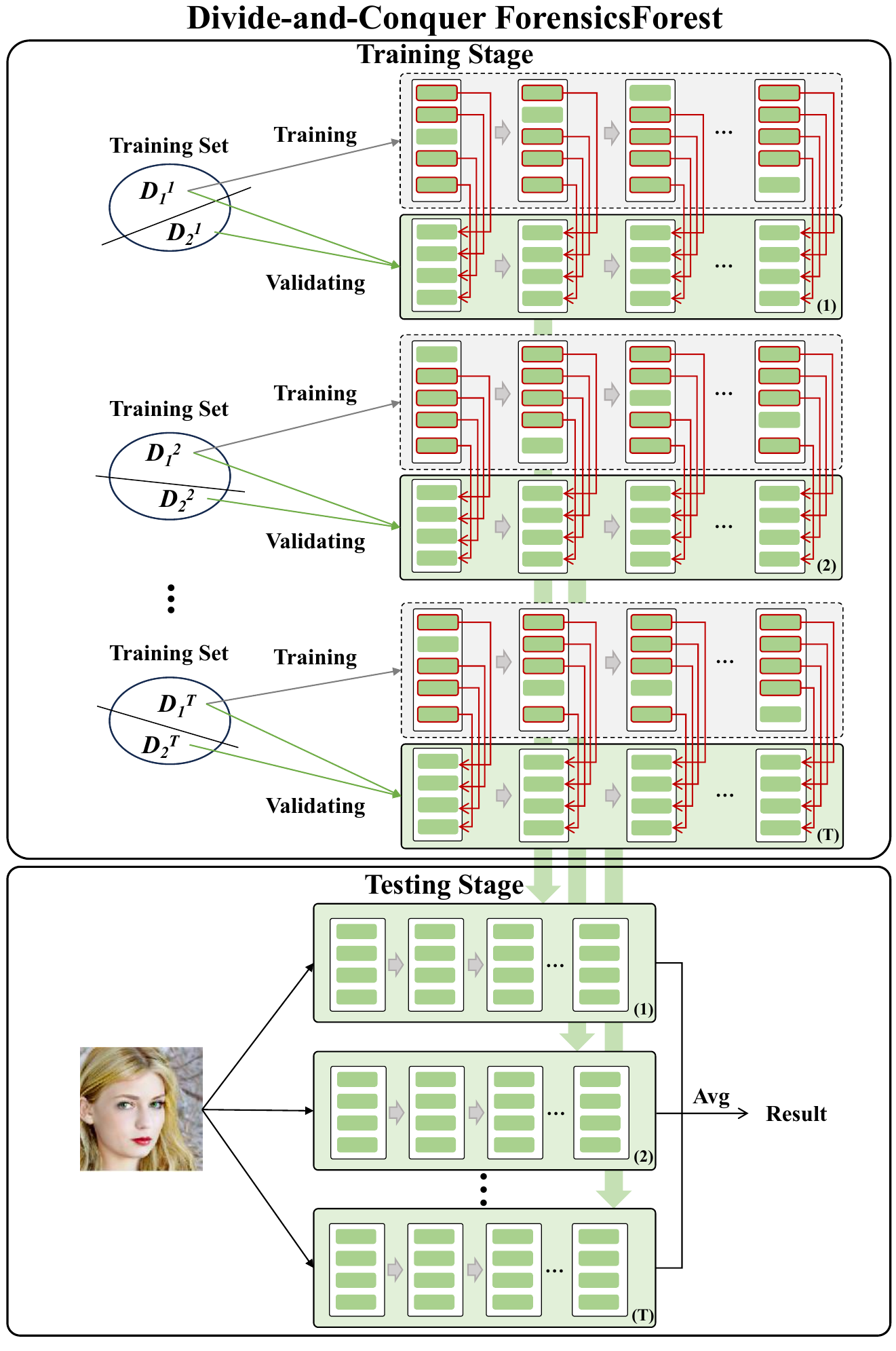}
\vspace{-0.25cm}
\caption{\small Overview of the training (top) and testing process (bottom) in Divide-and-Conquer ForensicsForest. In training, for each time, we use $\mc{D}_{1}^{i}$ to construct a primitive model and use $\mc{D}_{1}^{i}$ and $\mc{D}_{2}^{i}$ to select suitable components and assemble them as a final model. In testing, we use these assembled models and average the predictions as the final result. See text for details.}
\label{fig:divide}
\vspace{-0.3cm}
\end{figure}

Denote the whole training set as $\mc{D}$ and the repeated number as $T$. At the $i$-th time, we randomly divide  $\mc{D}$ into two portions, $\mc{D}_{1}^{i}$ and $\mc{D}_{2}^{i}$, using a ratio $r=\frac{|\mc{D}_{1}^{i}|}{|\mc{D}|}$ to control the size of each portion. We design a candidate ForensicsForest with $m (m \geq 4)$ forests in a layer and train it using $\mc{D}_{1}^{i}$.  After training, we need to select suitable components, \ie, forests from the candidate model, and assemble them to form a final ForensicsForest. To take into account both fitting and generalization, we assess the effectiveness of each forest in a layer by computing its average accuracy on $\mc{D}_{1}^{i}$ and $\mc{D}_{2}^{i}$. Then we select top-$n$ forests in a layer having the best average accuracy and assemble these forests into a layer to construct a ForensicsForest at $i$-th time. To be consistent with our ForensicsForst configuration, $n$ is set to $4$ which contains two random forests and two completely random forests respectively. This process is repeated for $T$ times and constructs a candidate forest model for each time. Finally we ensemble these forest models and average the output of each model as the final prediction. By doing so, the memory load of constructing a ForensicsForest can be divided into several small pieces, which significantly reduces the memory cost by tuning with a ratio $r$. The procedure is illustrated in Fig.~\ref{fig:divide}.

\section{Experiments}
\label{sub:experiments}

\subsection{Experimental Settings}
\smallskip
\noindent{\bf Datasets.}
Our method is validated on three types of GAN-generate face images, which are StyleGAN \cite{Karras_2019_CVPR}, StyleGAN2 \cite{karras2020analyzing} and StyleGAN3 \cite{karras2021alias} respectively. The face quality is improved along with the version of StyleGAN increasing. The real face images are collected from the Flickr-Faces-HQ (FFHQ) dataset \cite{Karras_2019_CVPR}. For StyleGAN and StyleGAN2 dataset, we randomly select $5,000$ images from FFHQ and StyleGAN or StyleGAN2 faces respectively. StyleGAN3 only provides $867$ images. Thus we randomly select the same amount of real images from FFHQ to construct the StyleGAN3 dataset. The ratio of training and testing set is $8:2$ for all datasets and all images have the same resolution of $1024 \times 1024$.


\smallskip
\noindent{\bf Implementation Details.}
The proposed ForensicsForest is trained and tested only using an Intel Core-i5 12400F CPU and no data augmentation is used. The number of forests in a layer is set to $4$ and the number of trees in a forest is set to $100$. We use four scales as {$N=1,2,3,4$}, where $N=2,3$ denotes to vertically split the input image into $2$ and $3$ patches. In training, the maximum number of layers is set to $20$ and the number of layers for checking the performance improvement is set to $2$. The maximum training epoch for each CNN model is set to $20$. All CNN models are trained using a single Nvidia 2080ti GPU. For Hybrid ForensicsForest, we set the learning rate to 0.01, and other basic settings are the same as ForensicsForest. For Divide-and-Conquer ForensicsForest, the ratio $r$ of training and validating is set to $0.9$, the repeated time number $T$ is set to $5$, and the number $m$ of forests in each candidate forest model is set to $16$.

\subsection{Results of ForensicsForest}
\label{sec:result-fforest}
\smallskip
\noindent{\bf Compared to CNN models.} To demonstrate the efficacy of our method, we compare our method with several mainstream CNN models, which are VGG16 \cite{simonyan2014very}, ResNet18 \cite{he2016deep}, ResNet50\cite{he2016deep}, Inception V3 \cite{szegedy2016rethinking}, MobileNet \cite{howard2017mobilenets}, ResNeXt \cite{xie2017aggregated}, MNASNet \cite{tan2019mnasnet}, EfficientNet \cite{tan2019efficientnet}, and RegNet \cite{radosavovic2020designing} respectively. We load their pre-trained weights on ImageNet and fine-tune all layers of them on each StyleGAN dataset. The learning rate is set to $0.001$.
The performance of our method in comparison to CNN models is shown in Table~\ref{tab:comparison} (Top).
The evaluation metrics are Accuracy (ACC) and Area Under Curve (AUC) following previous works. We can see for all of these datasets, our method achieves competitive and even better performance than CNN models. In particular, our method outperforms all CNN models by at least $4.7\%$ in ACC on the StyleGAN dataset. For example, our method surpasses ResNet50 by $6.8\%$ in ACC and outperforms MNASNet by a large margin, $13.3\%$, in ACC. Our method also performs very well on the StyleGAN2 dataset, which achieves $98.7\%$ in ACC and $99.8\%$ in AUC. StyleGAN3 dataset is more challenging as the synthesis quality is the best and the training set is small. Thus all methods perform compromised compared to other datasets. But it can be seen our method can still achieve the best performance on this dataset, which outperforms others by around $15\%$ in ACC and $11.3\%$ in AUC on average, demonstrating the effectiveness of our method on detecting GAN-generated face images.

\begin{table}[!t]
    \renewcommand{\arraystretch}{1.1}
    \centering

    \caption{\small Performance ($\%$) of different methods on three datasets.}
    \vspace{-0.2cm}
    \label{tab:comparison}
    \setlength{\tabcolsep}{1.5mm}{

    \begin{tabular}{c|cc|cc|cc}
    \hline
    \multirow{2}{*}{Method}     & \multicolumn{2}{c|}{\textbf{StyleGAN}}     & \multicolumn{2}{c|}{\textbf{StyleGAN2}}    & \multicolumn{2}{c}{\textbf{StyleGAN3}} \\ 
    \cline{2-7}     
    & ACC & AUC & ACC & AUC & ACC & AUC \\
    \hline
    \textbf{VGG16}\cite{simonyan2014very} & 93.5 & 99.7 & 97.8 & 100.0 & 72.3 & 86.4  \\ 
    \hline
    \textbf{ResNet18}\cite{he2016deep} & 92.2 & 98.9 & 86.1 & 96.1 & 70.2 & 85.3  \\ 
    \hline
    \textbf{ResNet50}\cite{he2016deep} & 93.2 & 99.6 & 90.1 & 96.5 & 81.5 & 89.8  \\ 
    \hline
    \textbf{Inception V3}\cite{szegedy2016rethinking} & 94.6 & 99.2 & 97.8 & 99.7 & 77.5 & 86.6  \\ 
    \hline
    \textbf{MobileNet}\cite{howard2017mobilenets} & 94.6 & 98.3 & 94.6 & 98.8 & 71.7 & 81.6  \\ 
    \hline
    \textbf{ResNeXt}\cite{xie2017aggregated} & 95.3 & 99.0 & 86.1 & 94.8 & 74.0 & 82.2  \\ 
    \hline
    \textbf{MNASNet}\cite{tan2019mnasnet} & 86.7 & 97.0 & 81.1 & 97.0 & 64.7 & 71.1  \\ 
    \hline
    \textbf{EfficientNet}\cite{tan2019efficientnet} & 90.3 & 96.9 & 92.1 & 97.7 & 74.1 & 89.3  \\ 
    \hline
    \textbf{RegNet}\cite{radosavovic2020designing} & 92.2 & 99.7 & 93.4 & 99.7 & 74.6 & 85.7  \\ 
    \hline 
    \hline
    
    \textbf{CNNDetection} \cite{wang2020cnn} & 98.4 & 99.8 & 97.9 & 99.7 & 78.9 & 95.1  \\ 
    \hline
    \textbf{GramNet} \cite{liu2020global} & 99.6 & 100.0 & 97.6 & 99.8 & 70.0 & 89.7  \\ 
    \hline
    \textbf{CoALBP+LPQ}\cite{boulkenafet2016face} & 93.8 & 98.3 & 97.4 & 99.6 & 60.3 & 64.4 \\
    \hline
    \textbf{Patch-forensics} \cite{patchforensics} & 98.1 & 99.7 & \textbf{99.4} & \textbf{100.0} & 62.9 & 71.7 \\
    \hline
    \textbf{AutoGAN}\cite{zhang2019detecting} & 93.9 & 99.2 & 92.2 & 98.1 & 77.7 & 93.5 \\
    \hline
    \hline 
    \textbf{Meso4} \cite{afchar2018mesonet} & 80.3 & 93.0 & 60.5 & 82.6 & 53.5 & 56.3  \\ 
    \hline
    \textbf{MesoInception} \cite{afchar2018mesonet} & 86.0 & 96.8 & 91.4 & 98.3 & 63.5 & 80.0  \\ 
    \hline
    \textbf{Xception} \cite{rossler2019faceforensics++} & 97.5 & 99.9 & 98.0 & 99.9 & 76.8 & 84.7 \\
    \hline
    \textbf{EfficientB4} \cite{tan2019efficientnet} & 93.0 & 99.8 & 94.6 & 99.9 & 69.4 & 85.6 \\
    \hline
    \textbf{Capsule} \cite{nguyen2019capsule} & 96.4 & 100.0 & 94.4 & 100.0 & 73.3 & 89.4 \\
    \hline
    \textbf{FWA} \cite{li2019exposing} & 98.5 & 100.0 & 98.8 & 100.0 & 79.4 & 87.0 \\
    \hline
    \textbf{Face X-ray} \cite{li2020face} & 97.6 & 99.9 & 97.8 & 100.0 & 75.7 & 88.5 \\
    \hline
    \textbf{FFD} \cite{dang2020detection} & 98.6 & 100.0 & 98.9 & 100.0 & 80.2 & 89.0 \\
    \hline
    \textbf{CORE} \cite{ni2022core} & 98.7 & 100.0 & 98.5 & 100.0 & 78.4 & 87.1 \\
    \hline
    \textbf{Recce} \cite{cao2022end} & 98.4 & 100.0 & 99.0 & 100.0 & 79.8 & 85.9 \\
    \hline
    \textbf{UCF} \cite{yan2023ucf} & 95.1 & 99.2 & 97.8 & 99.8 & 76.2 & 85.1 \\
    \hline
    \textbf{F3Net} \cite{qian2020thinking} & 98.5 & 100.0 & 99.0 & 100.0 & 80.2 & 89.4 \\
    \hline
    \textbf{SPSL} \cite{liu2021spatial} & 97.6 & 99.8 & 98.0 & 99.7 & 76.2 & 87.5 \\
    \hline
    \textbf{SRM} \cite{luo2021generalizing} & 98.7 & 100.0 & 98.7 & 100.0 & 77.6 & 87.6 \\
    \hline
    \hline
    \textbf{Ours} & \textbf{100.0} & \textbf{100.0} & 98.7 & 99.8 & \textbf{88.4} & \textbf{95.5}  \\ 
    \hline
    \end{tabular}}
    \vspace{-0.1cm}
\end{table}

\smallskip
\noindent{\bf Compared to GAN-generated Face Detection Methods.}
We also compare our method with five recent CNN-based GAN-generated face detection methods, CNNDetection \cite{wang2020cnn} and GramNet \cite{liu2020global}, {CoALBP+LPQ\cite{boulkenafet2016face}, Patch-forensics \cite{patchforensics} and AutoGAN\cite{zhang2019detecting}. For these methods, we retrain them on our datasets using their released codes. The results are shown in Table~\ref{tab:comparison} (Middle). We can observe that the compared five methods show good performance on the StyleGAN dataset. However, the performance of these methods drops on the StyleGAN3 dataset, especially at the ACC metric. In comparison, our method performs well and is stable in all datasets, which represents the effectiveness of our method. 

\smallskip
\noindent{\bf Compared to Deepfake Detection Methods.}
Besides the GAN-generated face detection methods, we also compare our method with recent Deepfake detection methods, including Meso4 \cite{afchar2018mesonet}, MesoInception\cite{afchar2018mesonet}, Xception\cite{rossler2019faceforensics++}, EfficientB4\cite{tan2019efficientnet}, Capsule\cite{nguyen2019capsule}, FWA\cite{li2019exposing}, Face X-ray\cite{li2020face}, FFD\cite{dang2020detection}, CORE\cite{ni2022core}, {Recce} \cite{cao2022end}, UCF\cite{yan2023ucf}, F3Net\cite{qian2020thinking}, SPSL\cite{liu2021spatial}, SRM\cite{luo2021generalizing}. For a fair comparison, we retrain these methods on our datasets following the codes provided in \cite{yan2023deepfakebench}. As shown in Table \ref{tab:comparison} (Bottom), most of the methods perform well on StyleGAN and StyleGAN2 datasets except Meso4. However, when tested on the StyleGAN3 dataset, there is a notable performance drop. This trend is consistent with GAN-generated face detection methods in general, as these methods mainly focus on detecting Deepfake forgery types, that typically manipulate the local area rather than the whole image in the GAN-generated faces. In contrast, our method outperforms these Deepfake detection methods by a large margin, particularly on the StyleGAN3 dataset.

\smallskip
\noindent{\bf Training Complexity.}
Table \ref{tab:time} records the time (seconds) of training corresponding methods. We compare our method with three classical architectures: VGG16, ResNet50, and Inception V3. This comparison is considered representative as these models are commonly used in recent counterpart methods, \eg, \cite{wang2020cnn,li2019exposing,dang2020detection,yan2023ucf}. It can be seen that despite CNN models being trained on GPU while our method is trained on CPU, our method has significantly lower time consumption than CNN models. 

\begin{table}[!t]
    \renewcommand{\arraystretch}{1.1}
    \centering
    \caption{\small Training time (seconds) of different methods.}
    \vspace{-0.2cm}
    \label{tab:time}
    \begin{tabular}{c|r|r|r}
    \hline
    {Method} & \textbf{StyleGAN} & \textbf{StyleGAN2} & \textbf{StyleGAN3} \\
    \hline
    \textbf{VGG16}\cite{simonyan2014very} & 3390.0s & 3486.8s & 229.5s \\
    \hline
    \textbf{ResNet50}\cite{he2016deep} & 2879.1s & 2646.7s & 212.6s \\
    \hline
    \textbf{Inception V3}\cite{szegedy2016rethinking} & 3745.3s & 3540.0s & 350.6s \\
    \hline
    {\textbf{Ours}} & \textbf{232.5s} & \textbf{224.1s} & \textbf{142.6s} \\
    \hline
    \end{tabular}
    \vspace{-0.1cm}
\end{table}

\smallskip
\noindent{\bf Performance on More Generative Models.}
We also validate our method on other generative models, such as ProGAN\cite{karras2018progressive}, StarGAN\cite{choi2018stargan}, and a recent diffusion-based model named LDM \cite{rombach2022high}. To construct these datasets, we run their official code under default settings. We generate $500$ face images as a fake set and select the exact amount of images from the FFHQ dataset as a real set. The training and testing split is set to $8:2$. The resolution of all images is $256 \times 256$. Table~\ref{tab:other_face} shows the performance of our method and other CNN models to detect generated faces. It can be seen that our method achieves the best performance, almost $100\%$ on ACC and $100\%$ on AUC, surpassing all other CNN models. This result demonstrates that our method is effective in exposing the fake faces generated by other models.

\begin{table}[!t]
    \renewcommand{\arraystretch}{1.1}
    \centering
    \caption{\small Performance ($\%$) of different methods on faces generated by ProGAN, StarGAN, and LDM.}
    \vspace{-0.2cm}
    \label{tab:other_face}
    \begin{tabular}{c|cc|cc|cc}
    \hline
    \multirow{2}{*}{Method}     & \multicolumn{2}{c|}{\textbf{ProGAN}}     & \multicolumn{2}{c|}{\textbf{StarGAN}}    & \multicolumn{2}{c}{\textbf{LDM}} \\ 
    \cline{2-7}     
    & ACC & AUC & ACC & AUC & ACC & AUC \\
    \hline
    \textbf{VGG16}\cite{simonyan2014very} & \textbf{99.0} & 99.9 & 96.0 & 99.6 & 99.5 & 100.0 \\
    \hline
    \textbf{ResNet50}\cite{he2016deep} & 97.0 & 99.5 & 93.0 & 97.5 & 99.5 & 100.0 \\
    \hline
    \textbf{Inception V3}\cite{szegedy2016rethinking} & 87.5 & 96.0 & 95.5 & 99.9  & 90.8 & 99.3 \\
    \hline
    {\textbf{Ours}} & 97.5 & \textbf{100.0} & \textbf{100.0} &\textbf{100.0} & \textbf{100.0} & \textbf{100.0} \\
    \hline
    \end{tabular}
    \vspace{-0.2cm}
\end{table}

\begin{table*}[!th]
    \centering
    \caption{\small {Performance of different methods under cross-dataset setting.}}
    \vspace{-0.2cm}
    \label{tab:cross}
    
    \begin{tabular}{c|c|cc|cc|cc|cc|cc|cc}
    \hline
    \multirow{2}{*}{Method}  & \multirow{2}{*}{Training} & \multicolumn{2}{c|}{\textbf{StyleGAN}}     & \multicolumn{2}{c|}{\textbf{StyleGAN2}} & \multicolumn{2}{c|}{\textbf{StyleGAN3}} & \multicolumn{2}{c|}{\textbf{ProGAN}}     & \multicolumn{2}{c|}{\textbf{StarGAN}} & \multicolumn{2}{c}{\textbf{LDM}} \\ 
    \cline{3-14}     
    & & ACC & AUC & ACC & AUC & ACC & AUC & ACC & AUC & ACC & AUC & ACC & AUC\\
    \hline
    \textbf{VGG16}\cite{simonyan2014very} & \multirow{7}{*}{\textbf{StyleGAN}} & 93.5 & 99.7 & 80.1 & 92.2 & 50.6 & 59.3 & 50.5 & 58.4 & 50.1 & 68.2 & 52.3 & 56.7 \\ 
    \cline{1-1} \cline{3-14}
    \textbf{ResNet50}\cite{he2016deep} & & 93.2 & 99.6 & 74.6 & 88.6 & 51.7 & 63.6 & \textbf{51.3} & 66.7 & 51.1 & 63.2 & 52.5 & 68.8 \\ 
    \cline{1-1} \cline{3-14}
    \textbf{Inception V3}\cite{szegedy2016rethinking} & & 94.6 & 99.2 & 68.3 & 84.7 & 53.8 & 65.2 & 50.9 & 66.7 & 50.3 & 56.1 & 51.5 & 58.4 \\ 
    \cline{1-1} \cline{3-14}
    \textbf{GramNet} \cite{liu2020global} & & 99.6 & 100.0 & 53.4 & 96.2 & 50.0 & \textbf{78.8} & 50.5 & \textbf{93.1} & 50.0 & 79.4 & 50.0 & 63.4  \\ 
    \cline{1-1} \cline{3-14}
    \textbf{Patch-forensics}\cite{patchforensics} & & 98.1 & 99.7 & 58.1 & 96.0 & 50.5 & 75.0 & 50.5 & 60.2 & \textbf{60.7}
    & 78.3 & 50.4 & 61.3   \\ 
    \cline{1-1} \cline{3-14}
    \textbf{AutoGAN}\cite{zhang2019detecting} & & 93.9 & 99.2 & 53.0 & 71.8 & 51.3 & 59.2 & 50.0 & 57.5 & 50.0 & 75.0 & 56.0 & 97.1  \\ 
    \cline{1-1} \cline{3-14}
    \textbf{Ours} & & \textbf{100.0} & \textbf{100.0} & \textbf{96.2} & \textbf{99.5} & \textbf{67.6} & 69.8 & 50.0 & 62.6 & 52.0 & \textbf{82.5} & \textbf{71.8} & \textbf{99.2} \\ 
    \hline 
    \hline
    \textbf{VGG16}\cite{simonyan2014very}  & \multirow{7}{*}{\textbf{StyleGAN2}} & 56.0 & 77.1 & 97.8 & 100.0 & 53.2 & 68.8 & 50.1 & 59.7 & 50.2 & 54.5 & 50.3 & 58.8 \\ 
    \cline{1-1} \cline{3-14}
    \textbf{ResNet50}\cite{he2016deep} & & 55.4 & 89.7 & 90.1 & 96.5 & 50.3 & 66.8 & 56.5 & 67.9 & 50.2 & 55.5 & 50.3 & 68.7\\
    \cline{1-1} \cline{3-14}
    \textbf{Inception V3}\cite{szegedy2016rethinking} & & 50.1 & 57.1 & 97.8 & 99.7 & 50.0 & 51.9 & 52.9 & 71.6 & 53.1 & 59.4 & 52.8 & 65.2\\ 
    \cline{1-1} \cline{3-14}
    \textbf{GramNet} \cite{liu2020global} & & 98.9 & 99.9 & 97.6 & 99.8 & 55.7 & 82.1 & \textbf{65.0} & \textbf{96.5} & 50.0 & 77.9 & 50.3 & 65.7 \\ 
    \cline{1-1} \cline{3-14}
    \textbf{Patch-forensics}\cite{patchforensics} & & 59.6 & 81.4 & \textbf{99.4} & \textbf{100.0} & 52.7 & 69.1 & 50.0 & 55.5 & 53.9 & 68.1 & 50.0 & 52.3 \\ 
    \cline{1-1} \cline{3-14}
    \textbf{AutoGAN}\cite{zhang2019detecting} & & 69.0 & 84.4 & 92.2 & 98.1 & \textbf{67.6} & \textbf{76.3} & 53.5 & 59.2 & 61.0 & 90.6 & \textbf{65.8} & 86.6 \\ 
    \cline{1-1} \cline{3-14}
    \textbf{Ours} & & \textbf{99.9} & \textbf{100.0} & 98.7 & 99.9 & 61.6 & 70.2 & 57.7 & 95.0 &  \textbf{80.9} & \textbf{98.8} & 62.5 & \textbf{98.9}\\ 
    \hline
    \hline
    \textbf{VGG16}\cite{simonyan2014very} & \multirow{7}{*}{\textbf{StyleGAN3}} & 69.8 & 76.2 & 80.8 & 89.0 & 72.3 & 86.4 & 65.0 & 70.6 & 56.5 & 57.5 & 59.5 & 61.7 \\ 
    \cline{1-1} \cline{3-14}
    \textbf{ResNet50}\cite{he2016deep} & & 59.9 & 72.1 & 65.8 & 83.9 & 81.5 & 89.8 & 61.2 & 70.1 & 54.2 & 58.6 & 58.8 & 63.1 \\ 
    \cline{1-1} \cline{3-14}
    \textbf{Inception V3}\cite{szegedy2016rethinking} & & 55.0 & 61.5 & 62.1 & 66.3 & 77.5 & 86.6 & 54.4 & 68.9 & 52.0 & 61.2 & 51.8 & 57.7\\ 
    \cline{1-1} \cline{3-14}
    \textbf{GramNet} \cite{liu2020global} & & \textbf{99.0} & \textbf{99.9} & \textbf{84.7} & \textbf{98.4} & 70.0 & 89.7 & \textbf{94.5} & \textbf{99.2} & 72.5 & 84.7 & 52.0 & 59.3 \\ 
    \cline{1-1} \cline{3-14}
    \textbf{Patch-forensics}\cite{patchforensics} & & 60.6 & 64.0 & 73.9 & 80.8 & 62.9 & 71.7 & 67.5 & 75.2 & 59.0 & 61.5 & 56.8 & 60.8 \\ 
    \cline{1-1} \cline{3-14}
    \textbf{AutoGAN}\cite{zhang2019detecting} & & 54.4 & 72.3 & 50.5 & 54.4 & 77.7 & 93.5 & 59.0 & 66.0 & 66.0 & 90.5 & 82.0 & 94.6 \\ 
    \cline{1-1} \cline{3-14}
    \textbf{Ours} & & 78.8 & 99.7 &  70.1 & 97.1 & \textbf{88.4} & \textbf{95.5} & 73.0 & 82.1 & \textbf{79.1} & \textbf{94.2} & \textbf{79.8} & \textbf{99.7}\\
    \hline
    \end{tabular}
    \vspace{-0.3cm}
\end{table*}

\smallskip
\noindent{\bf Performance on Cross-datasets.}
To rule out the bias in datasets, we conduct different cross-dataset experiments to validate the generalization ability of our method. Specifically, our method is trained on a dataset selected from StyleGAN, StyleGAN2, and StyleGAN3 and is tested on others. The results are shown in Table~\ref{tab:cross}. We can observe that our method generally performs better than other CNNs, especially in the setting of training on StyleGANs and testing on StarGAN and LDM. Moreover, compared to GAN-generated face detection methods, our method also shows favorable and competitive performance. We attribute it to the proposed hierarchical and multi-scale learning scheme in the forest model.
\smallskip
\noindent{\bf Ablation Study.}
This part studies the effect of various settings of ForensicsForest, including the input features, different multi-scale ensemble strategies, and their effects respectively.

\smallskip
\noindent{\em 1) Effect of Each Component of Input Features.}
Table \ref{tab:features} shows the effect of each component and their combinations of the input feature vector. The color histogram, power spectrum, and facial landmarks used in our method are denoted as CH, PS, and FL respectively. The first three rows represent using a color histogram, power spectrum, and facial landmarks respectively. The other rows represent combinations of these feature components in different ways. It can be seen that using an arbitrary component can achieve a decent performance on all datasets, affecting exposing the GAN-generated faces. Among these components, we can observe that for StyleGAN3, the frequency feature, \ie, power spectrum, performs best, and the biology feature, \ie, facial landmarks, achieves the second best. It is because frequency and biology features are usually not used in training the GAN models, thus the generated faces tend to lack these clues correspondingly. Using two combinations of these components can improve the performance than only using a single one, which represents each component is dedicated to one aspect, and can be complementary with other each. The last row is our method that considers all of these components, which achieves the best performance than others, demonstrating the effectiveness of each component in this task.

\begin{table}[t]
    \renewcommand{\arraystretch}{1.1}
    \centering
    \caption{\small Effect of different input features of our method.}
    \vspace{-0.2cm}
    \label{tab:features}
    \resizebox{\linewidth}{!}{
    \begin{tabular}{c|c|c|cc|cc|cc}
    \hline
    \multirow{2}{*}{CH} & \multirow{2}{*}{PS} & \multirow{2}{*}{FL}     & \multicolumn{2}{c|}{\textbf{StyleGAN}}     & \multicolumn{2}{c|}{\textbf{StyleGAN2}}    & \multicolumn{2}{c}{\textbf{StyleGAN3}} \\ 
    \cline{4-9}     
      & & & ACC & AUC & ACC & AUC & ACC & AUC \\
    \hline
    \checkmark & & & 94.0 & 98.5 & 96.3 & 99.2 & 69.7 & 77.9  \\ 
    \hline
    & \checkmark & & 99.8 & 100.0 & 96.8 & 99.8 & 73.4 & 86.0  \\ 
    \hline
    & & \checkmark & 91.3 & 97.0 & 92.4 & 97.4 & 71.7 & 79.8  \\ 
    \hline
    \checkmark & \checkmark & & 100.0 & 100.0 & 98.7 & 99.8 & 85.3 & 92.4 \\
    \hline
    & \checkmark & \checkmark & 100.0 & 100.0 & 96.8 & 99.8 & 78.0 & 94.8 \\
    \hline
    \checkmark & & \checkmark & 95.9 & 99.2 & 97.0 & 99.6 & 75.3 & 81.5 \\
    \hline
    \checkmark & \checkmark & \checkmark & \textbf{100.0} & \textbf{100.0} & \textbf{98.7} & \textbf{99.8} & \textbf{88.4} & \textbf{95.5}  \\
    \hline
    \end{tabular}}
\end{table}

\smallskip
\noindent{\em 2) Effect of Multi-scale Ensemble.} To validate the effectiveness of a multi-scale ensemble, we conduct experiments under two settings, which are with multi-scale (w/ ME) and without multi-scale (w/o ME). Table \ref{tab:ME} shows the performance of each case on three datasets. For (w/o ME), $N=1$ denotes using the whole image as input, and $N=4$ denotes using four image patches as input. For (w/ ME), {$N=1$ denotes integrating the augmented features of $N=4$ patches into the input features of the whole image for prediction, and the setting is the same for $N=4$, that is integrating the augmented features of the whole image into the input features of $N=4$ patches for prediction.} We can observe that using either the whole image or the four patches of the image can not reach the best performance, as it overlooks either the local or the global information. Especially for StyleGAN3, by using a multi-scale ensemble, the performance is further improved by $0.9\%$ when $N=1$ and $3.7\%$ when $N=4$ in ACC, as it considers both global and local information, which demonstrates its effectiveness in our method. Note that the multi-scale setting in this experiment is not the one used in our method. It is only used to demonstrate a simple multi-scale ensemble can still improve the performance. 


\begin{table}[t]
    \renewcommand{\arraystretch}{1.1}
    \centering
    \caption{\small Effect of a multi-scale ensemble of our method.}
    \vspace{-0.2cm}
    \label{tab:ME}
    \setlength{\tabcolsep}{2mm}{
    \begin{tabular}{cc|cc|cc}
    \hline
    \multirow{2}{*}{Dataset} & \multirow{2}{*}{Method}     & \multicolumn{2}{c|}{\textbf{N=1}}     & \multicolumn{2}{c}{\textbf{N=4}}    \\ 
    \cline{3-6}  
    & &  ACC & AUC & ACC & AUC \\
    \hline
    \multirow{2}{*}{\bf StyleGAN} & \textbf{w/o ME} & 99.9 & 100.0 & 99.9 & 100.0 \\ 
    
    & \textbf{w/ ME} & \textbf{100.0} & \textbf{100.0} & \textbf{100.0} & \textbf{100.0} \\ 
    \hline
    \multirow{2}{*}{\bf StyleGAN2} & \textbf{w/o ME} & 98.6 & 99.7 & 98.0 & 99.6 \\ 
    
    & \textbf{w/ ME} & \textbf{98.7} & \textbf{99.8} & \textbf{98.7} & \textbf{99.8}  \\ 
    \hline
    \multirow{2}{*}{\bf StyleGAN3} & \textbf{w/o ME} & 87.0 & 95.1 & 82.1 & 92.1  \\ 
    
    & \textbf{w/ ME} & \textbf{87.9} & \textbf{95.5} & \textbf{85.8} & \textbf{93.2} \\ 
    \hline
    \end{tabular}
    }
    \vspace{-0.2cm}
\end{table}

\smallskip
\noindent{\em 3) Various Ensemble Schemes.}
This part studies the effect of various ensemble strategies. Fig.~\ref{fig:ensembletypes} illustrates four ensemble schemes: E1 is to cascade the augmented features from each scale in order; E2 is to cascade the augmented features from each scale in order and add the global feature of $N=1$ at the last ensemble; E3 is to cascade all augmented features from previous scales into the next scale; E4 is the ensemble scheme used in our method, which integrates the augmented features from other scales into the first scale. As shown in Table~\ref{tab:scheme}, all these schemes can perform well, even perfectly, on StyleGAN. For StyleGAN2 and StyleGAN3, E1 and E2 perform slightly better than others respectively. In our method, we select E4 as it achieves similar performance but can parallel train $N=2,3,4$ scales and fuse with $N=1$ finally, saving a large amount of time than other schemes.

\begin{figure}[t]
\centering
\includegraphics[width=0.6\linewidth]{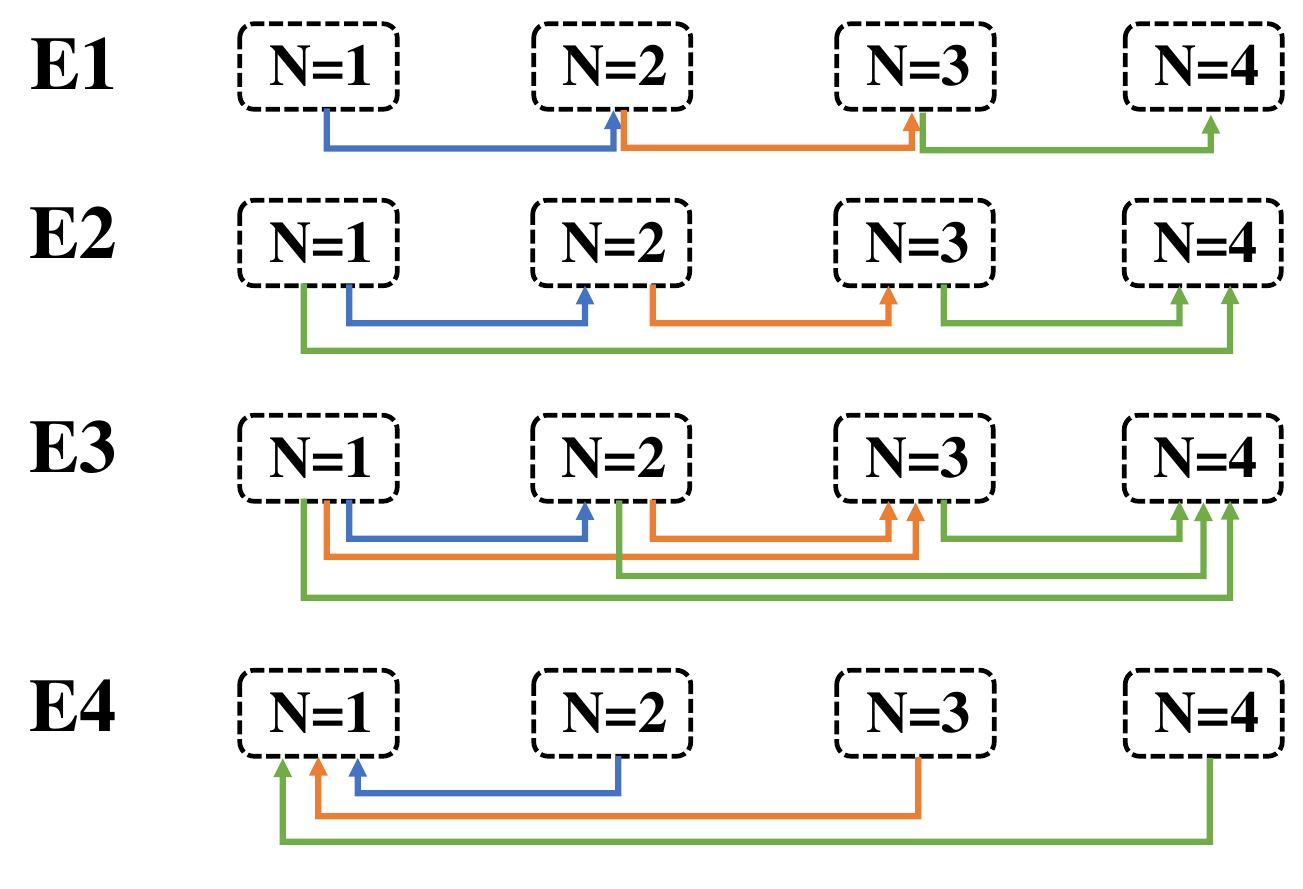}
\vspace{-0.25cm}
\caption{\small Illustration of different ensemble schemes.}
\label{fig:ensembletypes}
\end{figure}

\begin{table}[t]
    \renewcommand{\arraystretch}{1.1}
    \centering
    \caption{\small Effect of different multi-scale ensemble schemes.}
    \vspace{-0.2cm}
    \label{tab:scheme}
    \begin{tabular}{c|cc|cc|cc}
    \hline
    \multirow{2}{*}{Ensemble}     & \multicolumn{2}{c|}{\textbf{StyleGAN}}     & \multicolumn{2}{c|}{\textbf{StyleGAN2}}    & \multicolumn{2}{c}{\textbf{StyleGAN3}} \\ 
    \cline{2-7}     
    & ACC & AUC & ACC & AUC & ACC & AUC \\
    \hline
    \textbf{E1} & 99.8 & 99.9 & \textbf{99.3} & \textbf{99.9} & 88.7 & 95.4  \\ 
    \hline
    \textbf{E2} & 99.9 & 100.0 & 98.8 & 99.9 & \textbf{89.9} & \textbf{96.2} \\
    \hline 
    \textbf{E3} & 99.8 & 100.0 & 98.9 & 99.8 & 89.9 & 95.2  \\ 
    \hline
    \textbf{E4} & \textbf{100.0} & \textbf{100.0} & 98.7 & 99.8 & 88.4 & 95.5  \\ 
    \hline
    \end{tabular}
    \vspace{-0.1cm}
\end{table}

\smallskip
\noindent{\bf Robustness.}
This part studies the robustness of our method against resizing, JPEG compression, brightness change, and additive noises, respectively. 

\smallskip
\noindent{\em 1) {Resizing.}} 
This part studies the performance of our method by testing on different image resolutions. Note that our method is independent of the image size, thus it can be applied to arbitrary sizes of images. Specifically, the images are resized to four types, $256\times256$, $512\times512$, $768\times768$, $1024 \times 1024$ respectively. As shown in Fig.~\ref{fig:robust} (the first column), the performance drops a lot when the resolution becomes small compared to CNN models on StyleGAN and StyleGAN2 datasets. But for the StyleGAN3 dataset, all methods drop significantly and our method achieves a similar trend to other methods. More analyses about the effect of image sizes are conducted in Section~\ref{subsec:limits}.

\noindent{\em 2) JPEG Compression.} We use OpenCV to change the compression level of input images, ranging from $20$ to $100$. The larger value denotes less compression, and $100$ means no compression is applied. Fig.~\ref{fig:robust} (the second column)shows the performance of our method and CNN-based methods against JPEG compression. Note these methods are trained on regular images and tested on compressed images. A similar trend is observed in these figures: all these methods drop the performance with the compression level increasing. Benefiting from a large number of parameters in CNNs, most CNN-based methods perform better than ours, \ie, they drop more slowly compared to ours. Nevertheless, our method can still achieve decent performance even at the highest compression level. We leave the improvement of the robustness of our method to future work. 


\smallskip
\noindent{\em 3) Brightness Change.} Since we utilize color histograms as the feature for detection, and the color histogram is usually sensitive to light, we study the robustness of our method against brightness changes. We also use OpenCV to change the brightness of images. Specially, $1$ indicates no change, and $<1$ indicates brighter, while $>1$ indicates darker.  Fig.~\ref{fig:robust} (the third column) shows the curve of our method facing different brightness factors, which represents that our method can resist a certain brightness change compared to CNN models. Especially, on StyleGAN and StyleGAN3, our method performs best when the images become dark.


\smallskip
\noindent{\em 4) {Additive Noise.}}
For further evaluation, we add different levels of Gaussian noises to the images, making the standard deviation from $0$ to $5$. Note $0$ means no noise is added. Fig.~\ref{fig:robust} (the last column) shows a similar curve, revealing that our method can resist certain noises on more challenging datasets compared to CNN models.

\begin{figure}[t]
\centering
\includegraphics[width=\linewidth]{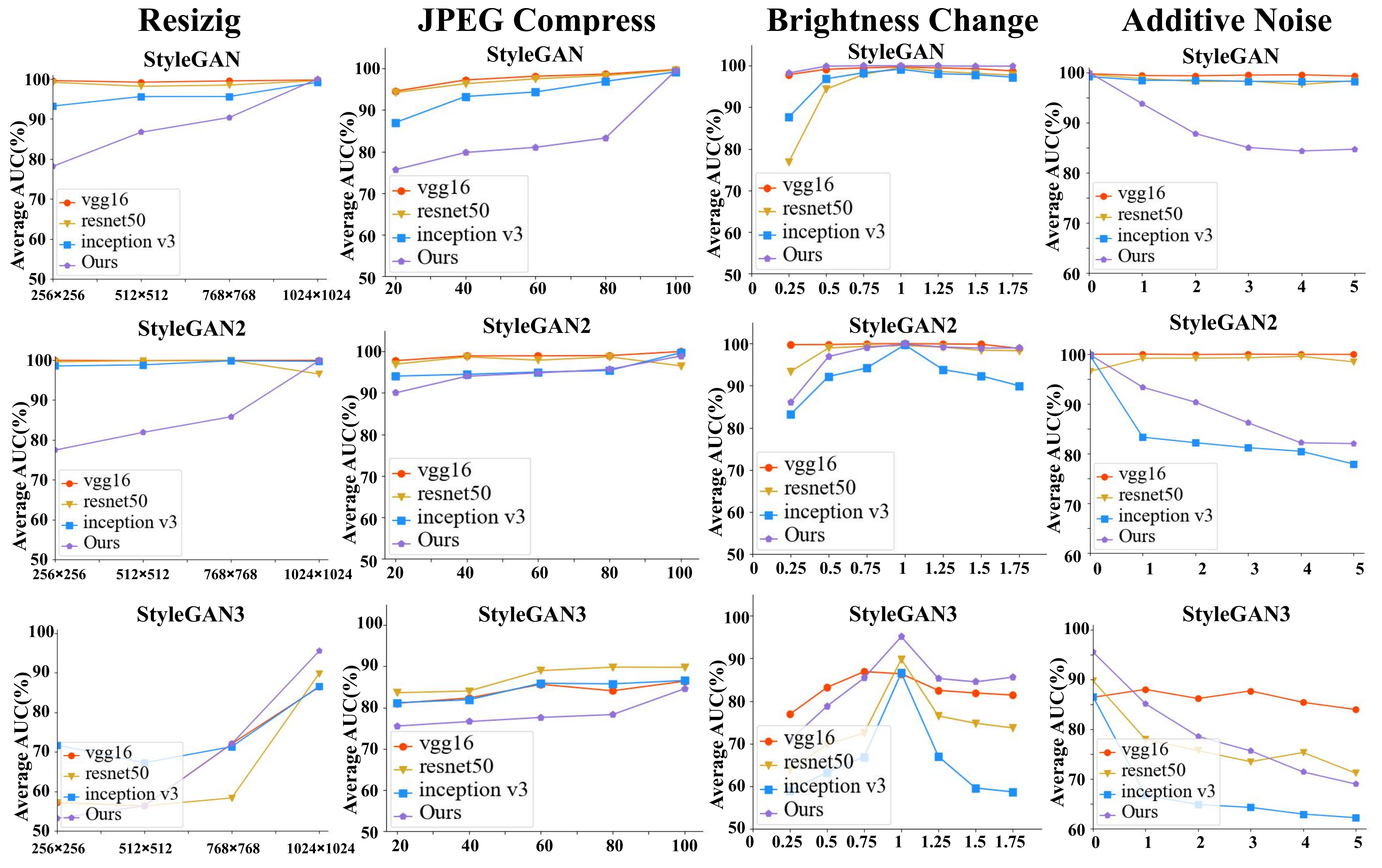}
\vspace{-0.6cm}
\caption{\small Performance of different methods against resizing, JPEG compression, brightness change, and additive noise.}
\label{fig:robust}
\end{figure}

\subsection{Results of Hybrid ForensicsForest}
\label{sub:hybridff}
Table~\ref{tab:FF_series} shows the performance of ForensicsForest and Hybrid ForensicsForest, respectively. ``FF'' stands for ForensicsForest. It can be seen that after integrating the FC layer into ForensicsForest, it performs notably better on challenging StyleGAN3 datasets, improving ACC and AUC scores by $3.2\%$ and $0.9\%$, respectively, demonstrating the effectiveness of added FC layers for augmented features refinement.
\begin{table}[t]
    \centering
    \renewcommand{\arraystretch}{1.1}
    \caption{\small Performance ($\%$) of ForensicsForest, and Hybrid ForensicsForest, respectively. ``FF'' stands for ForensicsForest. }
    \vspace{-0.2cm}
    \label{tab:FF_series}
    \setlength{\tabcolsep}{1.5mm}{
    \begin{tabular}{c|cc|cc|cc}
    \hline
    \multirow{2}{*}{Method}     & \multicolumn{2}{c|}{\textbf{StyleGAN}}     & \multicolumn{2}{c|}{\textbf{StyleGAN2}}    & \multicolumn{2}{c}{\textbf{StyleGAN3}} \\ 
    \cline{2-7}     
    & ACC & AUC & ACC & AUC & ACC & AUC \\
    \hline
    \textbf{FF} & \textbf{100.0} & \textbf{100.0} & 98.7 & 99.8 & 88.4 & 95.5  \\
    \hline
    \textbf{Hybrid FF} & 99.9 & 100.0 & \textbf{98.8} & \textbf{99.9} & \textbf{91.6} & \textbf{96.4}  \\ 
    \hline
    \end{tabular}
    }
    \vspace{-0.2cm}
\end{table}

\smallskip
\noindent{\bf Ablation Study.} In this part, we investigate several settings of adding FC layers into ForensicsForest, including the integration strategies and the effect of the scale of FC layers. 

\smallskip
\noindent{\em 1) Effect of Various Integration Strategies.}
Specifically, we study three settings: (1) we only add a FC layer at the end of our ForensicsForest (denoted as Hybrid1), (2) we simply add a FC layer after each forest layer, and send the output of FC layer into next forest layer (denoted as Hybrid2), (3) we stack the augmented features from previous forest layer to the next forest layer, as the input of FC layer in next forest layer (denoted as Hybrid3). The difference between these settings is illustrated in Fig.~\ref{fig:different-hybrid}. 
Table~\ref{tab:different_fc} shows the results of different settings, which reveals that Hybrid3 performs best compared to the other two settings, as it considers the knowledge from previous layers, offering sufficient information for the FC layer to precisely refine the augmented features from the forest layer. Thus we employ the third setting in our method.

\begin{figure}[t]
\centering
\includegraphics[width=0.95\linewidth]{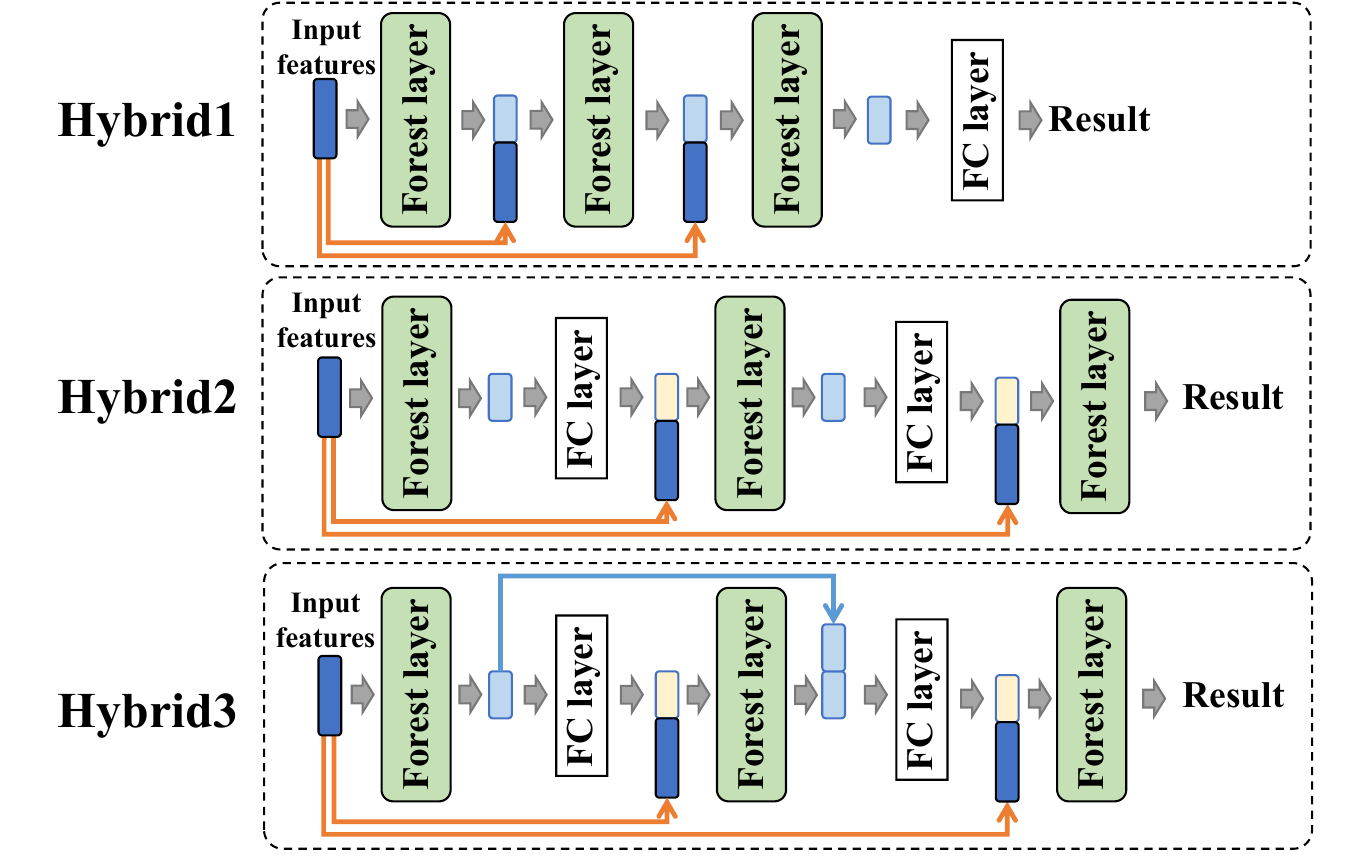}
\vspace{-0.25cm}
\caption{\small Illustration of various integration strategies in Hybrid ForensicsForest.}
\label{fig:different-hybrid}
\end{figure}

\begin{table}[t]
    \centering
    \renewcommand{\arraystretch}{1.1}
    \caption{\small Performance ($\%$) of different integration strategies in Hybrid ForensicsForest.}
    \vspace{-0.2cm}
    \label{tab:different_fc}
    \setlength{\tabcolsep}{1.5mm}{
    \begin{tabular}{c|cc|cc|cc}
    \hline
    \multirow{2}{*}{Method}     & \multicolumn{2}{c|}{\textbf{StyleGAN}}     & \multicolumn{2}{c|}{\textbf{StyleGAN2}}    & \multicolumn{2}{c}{\textbf{StyleGAN3}} \\ 
    \cline{2-7}     
    & ACC & AUC & ACC & AUC & ACC & AUC \\
    \hline
    \textbf{Hybrid1} & 99.9 & 100.0 & 98.3 & 99.8 & 85.4 & 94.7  \\ 
    \hline
    \textbf{Hybrid2} & 99.8 & 100.0 & 98.6 & 99.9 & 88.2 & 95.0  \\ 
    \hline
    \textbf{Hybrid3} & \textbf{99.9} & \textbf{100.0} & \textbf{98.8} & \textbf{99.9} & \textbf{91.6} & \textbf{96.4}  \\ 
    \hline
    \end{tabular}
    }
    \vspace{-0.2cm}
\end{table}

\smallskip
\noindent{\em 2) Effect of the Scale of FC layer.}
This part investigates the effect of different parameters in FC layer. For better understanding, for example, we use ``$a-b$'' to represent adding two FC layers and these layers have $a, b$ parameters respectively. Note that $a, b$ are the output dimensions of FC layers and we use them to represent the scale for simplicity. Table~\ref{tab:different_parameters} shows several settings of FC layers. The first row is the setting used in our method, in which we use one FC layer with 8 parameters. We also study the effect of using two FC layers with more parameters, \eg, $16-8$,$32-8$,$64-8$,$128-8$. It can be seen with the number of parameters increases, the performance is not improved as we expect. On the contrary, the first row with fewer parameters performs better than the others. We conjecture that in our integration, the efficacy of augmented features from the forest layer plays an important role in the final result. Despite we can employ FC layers for refinement, it may have an upper bound for the performance no matter what scale of FC layers you use. 

\begin{table}[t]
    \centering
    \renewcommand{\arraystretch}{1.1}
    \caption{\small Performance ($\%$) of Hybrid ForensicsForest with different number of parameters.}
    \vspace{-0.2cm}
    \label{tab:different_parameters}
    \setlength{\tabcolsep}{1.5mm}{
    \begin{tabular}{c|cc|cc|cc}
    \hline
    \multirow{2}{*}{Method}     & \multicolumn{2}{c|}{\textbf{StyleGAN}}     & \multicolumn{2}{c|}{\textbf{StyleGAN2}}    & \multicolumn{2}{c}{\textbf{StyleGAN3}} \\ 
    \cline{2-7}     
    & ACC & AUC & ACC & AUC & ACC & AUC \\
    \hline
    \textbf{8} & 99.9 & \textbf{100.0} & \textbf{98.8} & \textbf{99.9} & \textbf{91.6} & \textbf{96.4}  \\ 
    \hline
    \textbf{16-8} & \textbf{100.0} & 100.0 & 98.6 & 99.9 & 88.2 & 95.5  \\ 
    \hline
    \textbf{32-8} & 100.0 & 100.0 & 98.4 & 99.9 & 89.9 & 96.1  \\ 
    \hline
    \textbf{64-8} & 100.0 & 100.0 & 98.8 & 99.9 & 89.3 & 95.7  \\ 
    \hline
    \textbf{128-8} & 100.0 & 100.0 & 98.6 & 99.9 & 87.9 & 96.0  \\ 
    \hline
    \end{tabular}
    }
\end{table}

\begin{table}[!ht]
    \centering
    \renewcommand{\arraystretch}{1.1}
    \caption{\small Performance ($\%$) of ForensicsForest, and Divide-and-Conquer ForensicsForest respectively. ``FF'' stands for ForensicsForest. ``D-and-C'' stands for Divide-and-Conquer.}\vspace{-0.2cm}
    \label{tab:DC-FF}
    \setlength{\tabcolsep}{1.5mm}{
    \begin{tabular}{c|cc|cc|cc}
    \hline
    \multirow{2}{*}{Method}     & \multicolumn{2}{c|}{\textbf{StyleGAN}}     & \multicolumn{2}{c|}{\textbf{StyleGAN2}}    & \multicolumn{2}{c}{\textbf{StyleGAN3}} \\ 
    \cline{2-7}     
    & ACC & AUC & ACC & AUC & ACC & AUC \\
    \hline
    \textbf{FF} & 100.0 & 100.0 & \textbf{98.7} & 99.8 & \textbf{88.4} & 95.5  \\ 
    \hline
    \textbf{D-and-C FF} & \textbf{100.0} & \textbf{100.0} & 98.6 & \textbf{99.9} & 87.6 & \textbf{95.6}  \\ 
    \hline
    \end{tabular}
    }
    \vspace{-0.2cm}
\end{table}


\begin{figure}[t]
\centering
\includegraphics[width = \linewidth]{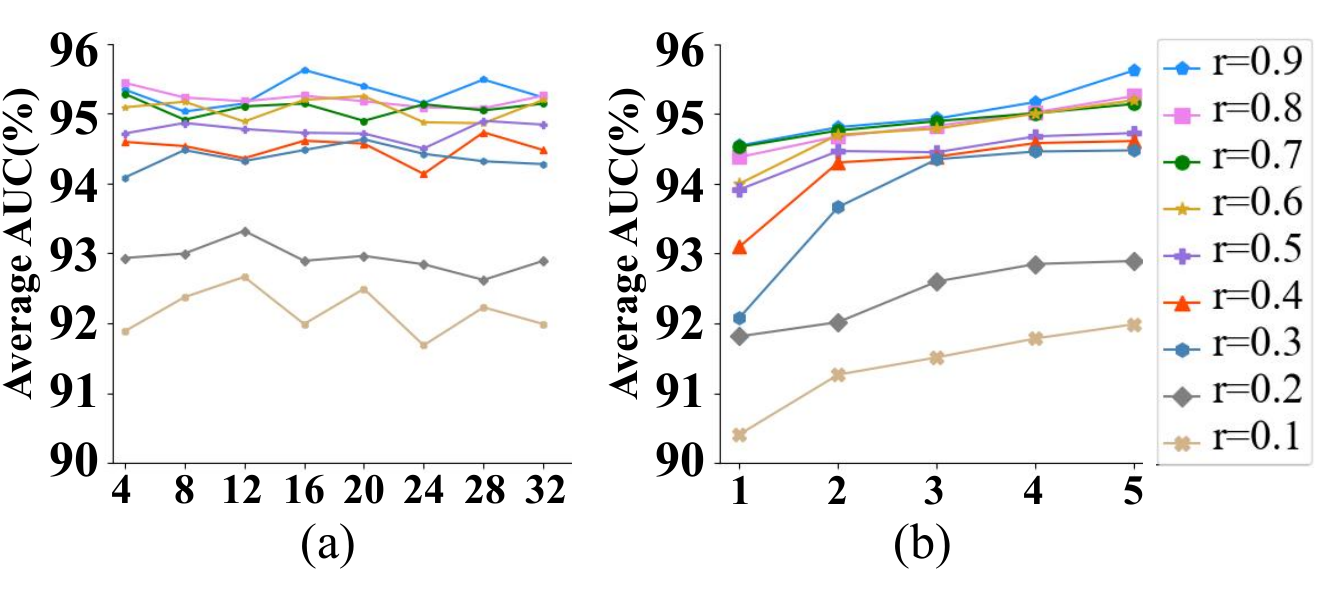}
\vspace{-0.7cm}
\caption{\small (a) Effect of different $r$, $m$ on Divide-and-Conquer ForensicsForest. (b) Effect of $T$ on Divide-and-Conquer ForensicsForest. The definition of $r,m,T$ can be found in Section \ref{sub:Divide-and-Conquer}.}
\label{fig:r-and-m}
\vspace{-0.2cm}
\end{figure}

\subsection{Results of Divide-and-Conquer ForensicsForest}
Table~\ref{tab:DC-FF} shows the performance of ForensicsForest and Divide-and-Conquer ForensicsForest, respectively. We use ``D-and-C FF'' to denote Divide-and-Conquer ForensicsForest. Our Divide-and-Conquer ForensicsForest can achieve competitive performance compared to our native ForensicsForest, which only slightly drops on the StyleGAN3 dataset by $0.8\%$ on ACC, demonstrating the effectiveness of the divide-and-conquer strategy to retain the performance while saving the memory cost.

\smallskip
\noindent{\bf Ablation Study.} We investigate the effect of Divide-and-Conquer ForensicsForest using various parameters, such as the number of forests constructed before selection, the ratio of training and validating set each time, and the number of repeated times, respectively.

\smallskip
\noindent{\em 1) Effect of Different Number $m$ of Forests before Selection and the Ratio $r$ of Training and Validating} Set.
We conduct many experiments with different $m$ and $r$, and the results are illustrated in Fig.~\ref{fig:r-and-m} (a). The x-axis is the number $m$ of forests before the selection, and different figures represent the performance of our method using different ratios $r$. Note that we always select four forests from $m$, thus the x-axis can only start from $m=4$, which means using all forests without selection. We study our method on eight different forest numbers $m$, which range from $4$ to $32$, and nine different ratios $r$, which range from $0.1$ to $0.9$. As shown in these figures, we can observe that the performance improves gradually with the ratio $r$ increasing. It is consistent with our expectation, as increasing the ratio $r$ can provide more samples for training, improving the performance subsequently. But we can also observe that the performance starts to become stable at a ratio $r=0.4$, which demonstrates the effectiveness of our Divide-and-Conquer ForensicsForest can learn sufficient knowledge by using only a small portion of training samples. For $m$, our preliminary expectation is that the performance should be improved by increasing $m$, as larger $m$ means providing more options for selection. However, the observation is inconsistent with our expectation, that the larger $m$ may not bring a performance boost than less $m$. We conjecture that it is highly related to $r$, as small $r$ means only using a small portion of training samples, that a small number of forests is sufficient to capture the full knowledge, but a large number of forests may be underfitting. With the $r$ increasing, we can see the performance is not sensitive with $m$, which represents the training samples can provide sufficient knowledge to our model even though the forest number is not large.

\smallskip
\noindent{\em 2) Effect of Different Number of Repeated Times $T$.} Fig.~\ref{fig:r-and-m} (b) illustrates the performance of our method under different repeated times $T$ with different ratio $r$. It can be seen that for all cases of ratio $r$, the trend is the same in that the performance improves with the repeated time $T$ increasing. This is because more repeated time denotes more ForensicsForest models, which can provide more experience for final prediction.
\subsection{Discussions}
\label{subsec:limits}

\smallskip
\noindent{\bf Why Our Method Works?}
Observing that the GAN-generated faces usually exhibit distribution differences in three aspects, that are appearance, frequency, and biology respectively. This is because the objective functions used for training generative models can hardly take these aspects into account, which consequently reflects on the generated faces. For example, when training generative models, a discriminator is typically employed to distinguish if a face image is generated or not. The learning of this discriminator relies on two sets of faces, the real faces and the generated faces, which automatically learns the difference between them in a discriminator perspective. However, there are no explicit restrictions on the frequency distribution in the learning process, resulting in the inconsistency in frequency distribution between real and fake faces, 
Based on these aspects, we design a specific model that can effectively utilize and refine the input features to expose the more distinguishable traces.}

\begin{figure}[t]
\centering
\includegraphics[width=\linewidth]{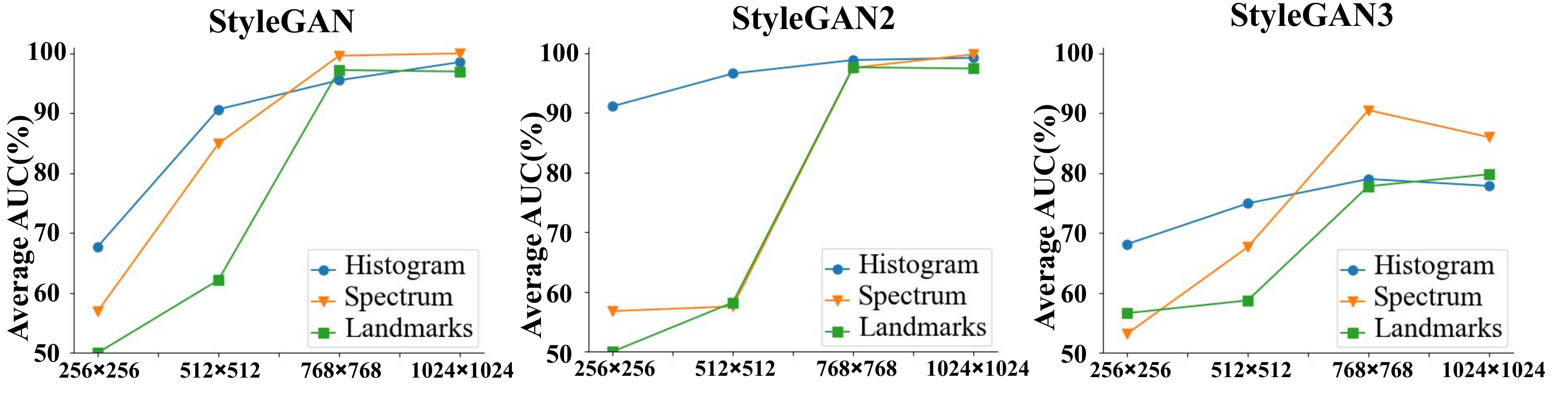}
\vspace{-0.6cm}
\caption{\small {Performance on images with different resolutions.}}
\label{fig:resize_analysis}
\vspace{0.1cm}
\end{figure}

\begin{figure}[ht]
\centering
\includegraphics[width=\linewidth]{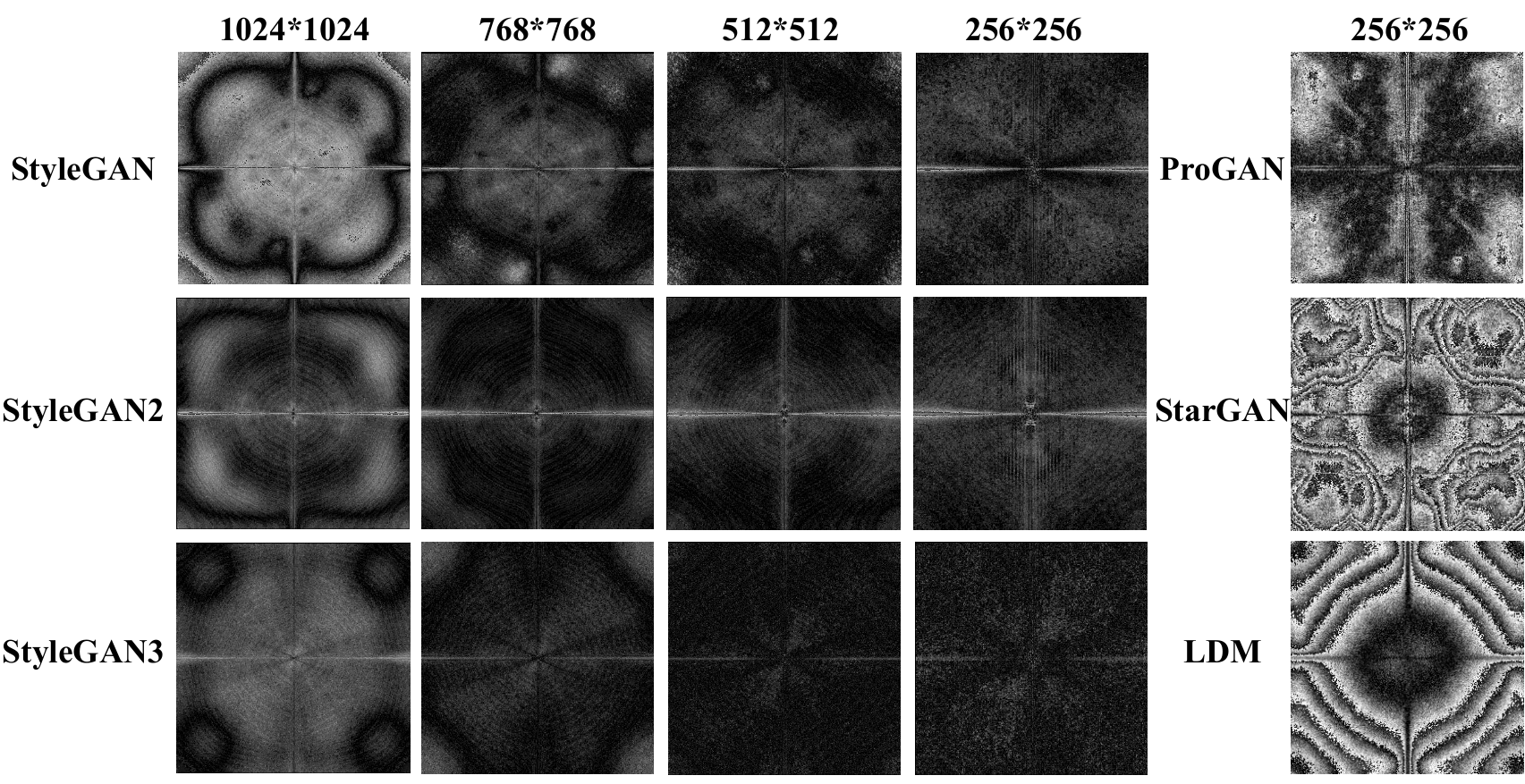}
\vspace{-0.6cm}
\caption{\small {Visualization of frequency difference maps of different datasets with various resolutions. The difference is multiplied by a factor of $40$ for a better view.}}
\label{fig:Frequency}
\vspace{-0.2cm}
\end{figure}

\smallskip
\noindent{\bf Possible Limitations.}
Recall the robustness analysis in Section~\ref{sec:result-fforest} that the small size (\eg, $256 \times 256$) may limit the performance of our method in StyleGAN compared to a larger size (\eg, $1024 \times 1024$). To figure out the reason behind this, we resize both the training and testing images to different resolutions and employ only one feature component for evaluation. The results are visualized in Fig.~\ref{fig:resize_analysis}. It can be seen that with the image size decreasing, the effectiveness of the frequency spectrum and landmark features are greatly disturbed compared to the color histogram, resulting in a clear performance drop. 
However, the results in Table~\ref{tab:other_face} are the opposite, where our method performs very well on ProGAN, StarGAN, and LDM, even if the image size is $256 \times 256$. For further analysis, we show the performance of using each feature component on these datasets in Table~\ref{tab:analysis}. It reveals that all the features are effective even in small image sizes on ProGAN and StarGAN faces. These results are reasonable as the quality of ProGAN and StarGAN faces does not catch with StyleGAN faces, thus these features still have distinguishable abilities even in small sizes. But for LDM, which is the latest generation model, we find a similar trend in StyleGAN, that the color histogram and facial landmark become less effective, only $88.0\%$ and $68.7\%$ at AUC respectively. However, compared to StyleGAN, the frequency spectrum in LDM is still highly effective, which is the key to retaining favorable performance. Thus, we conjecture that the frequency spectrum in LDM is more significant than StyleGAN. For validation, we visualize the frequency difference of the faces generated by different models in Fig.~\ref{fig:Frequency}. These maps are obtained by averaging the difference between fake and real faces. It can be seen that the frequency difference becomes inconspicuous with image size decreasing in StyleGAN datasets, while it is still apparent in ProGAN, StarGAN, and LDM. Thus, our method may not be robust when applied to the generative methods that have a frequency spectrum that is highly dependent on image sizes. 

\begin{table}[t]
    \centering
    \caption{\small {Effect of using different input features on different datasets.}}
    \vspace{-0.2cm}
    \label{tab:analysis}
    \begin{tabular}{c|c|c|cc|cc|cc}
    \hline
    \multirow{3}{*}{CH} & \multirow{3}{*}{PS} & \multirow{3}{*}{FL} & \multicolumn{2}{c|}{\textbf{ProGAN}} & \multicolumn{2}{c|}{\textbf{StarGAN}} & \multicolumn{2}{c}{\textbf{LDM}}  \\ 
       
    & & & \multicolumn{2}{c|}{\textbf{$256 \times 256$}}  & \multicolumn{2}{c|}{\textbf{$256 \times 256$}} & \multicolumn{2}{c}{\textbf{$256 \times 256$}} \\
    \cline{4-9}  
      & & & ACC & AUC & ACC & AUC & ACC & AUC \\
    \hline
    \checkmark & & & 82.0 & 92.3 & 100.0 & 100.0 & 80.8 & 88.0 \\ 
    \hline
    &  \checkmark & &96.5 & 99.9 & 100.0 & 100.0 & 100.0 & 100.0 \\ 
    \hline
    & & \checkmark & 93.5 & 98.2 & 99.0 & 99.0 & 63.5 & 68.7\\
    \hline
    \end{tabular}
    \vspace{-0.1cm}
\end{table}

\begin{figure}[!t]
\vspace{-0.5cm}
\centering
\includegraphics[width=.9\linewidth]{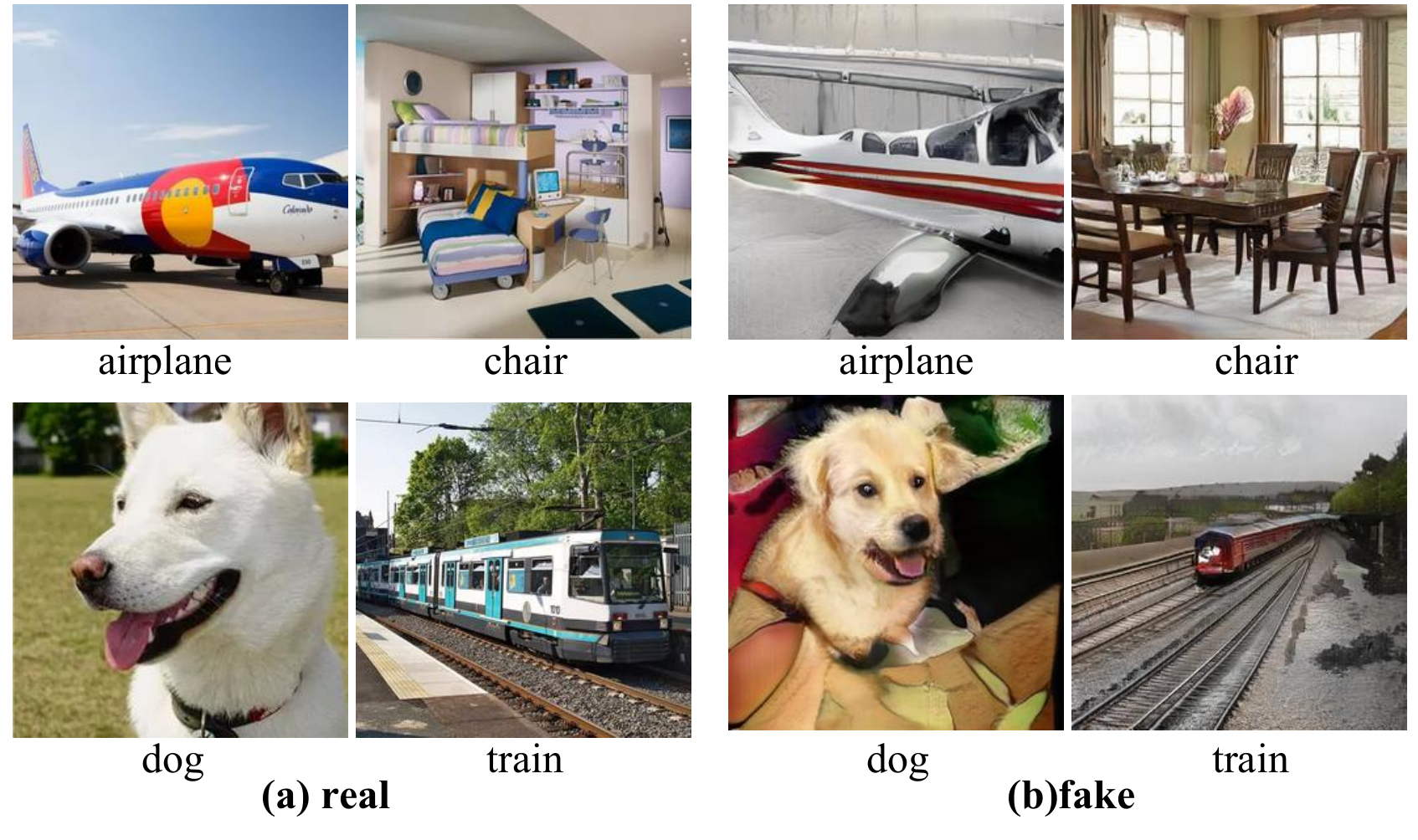}
\vspace{-0.3cm}
\caption{\small Examples of real images from LSUN dataset and generated images by ProGAN.}
\label{fig:proganexamples}
\end{figure}

\smallskip
\noindent{\bf Tentative Study on General Scenes.}
This part tentatively studies the feasibility of our method on detecting general GAN-generated images. Specifically, We select $10400$ real images from LSUN dataset \cite{yu2015lsun} and create the same number of images with general scenes using ProGAN \cite{karras2017progressive}. This dataset covers $20$ general classes and the resolution of all images is $256 \times 256$. Fig.~\ref{fig:proganexamples} shows several examples. Since faces are not contained in this scenario, we only use appearance and frequency features in detection. As shown in Table \ref{tab:progan_comparison}, our method can also achieve competitive performance.

\begin{table}[t]
    \centering
    \small
    \caption{\small Performance ($\%$) of different methods on ProGAN.}
    \vspace{-0.2cm}
    \label{tab:progan_comparison}
    \renewcommand{\arraystretch}{1.1}
    \setlength{\tabcolsep}{1.5mm}{
    \begin{tabular}{c|cc}
    \hline
    \multirow{2}{*}{Method}     & \multicolumn{2}{c}{\textbf{ProGAN}}  \\ 
    \cline{2-3}     
    & ACC & AUC \\
    \hline
    \textbf{VGG16}\cite{simonyan2014very} & 99.7 & 99.9  \\ 
    \hline
    \textbf{ResNet50}\cite{he2016deep} & 99.9 & 100.0 \\ 
    \hline
    \textbf{Inception V3}\cite{szegedy2016rethinking} & 99.9 & 100.0   \\ 
    \hline
    \textbf{Ours} & 93.7 & 98.5  \\ 
    \hline
    \end{tabular}
    }
    \vspace{-0.1cm}
\end{table}

\section{Conclusion}
This paper describes {\em ForensicsForest Family}, a novel set of forest-based methods to detect GAN-generate faces. In contrast to the recent efforts of using CNNs, we investigate the feasibility of using forest models and introduce a series of multi-scale hierarchical cascade forests, which are ForensicsForest, Hybrid ForensicsForest, and Divide-and-Conquer ForensicsForest, respectively. The ForensicsForest model contains three key components: input feature extraction, hierarchical cascade forest, and multi-scale ensemble. These components are specifically designed for GAN-generated face detection. Building on the ForensicsForest, we develop the Hybrid ForensicsForest that integrates CNN layers into the model to enhance the augmented features derived from forest layers. Moreover, to reduce the memory cost of training, we develop the Divide-and-Conquer ForensicsForest, which uses a randomly selected portion of training samples to construct models and pick up the suitable components to form a final forest model. Extensive experiments are conducted on multiple types of GAN-generated faces compared to recent CNN models and dedicated CNN-based detection methods, demonstrating that our method is surprisingly effective in exposing GAN-generated faces. 

\smallskip
\noindent{\bf Acknowledgement.} Yuezun Li was supported by the China Postdoctoral Science Foundation under Grant No. 2021TQ0314 and 2021M703036, the Fundamental Research Funds for the Central Universities. Bin Li was supported in part by NSFC (Grant U23B2022). Siwei Lyu was supported in part by NSF SaTC-2153112.
\vspace{-0.3cm}
\bibliographystyle{IEEEbib}
\bibliography{TIFS2023template}

\vspace{-1cm}

\begin{IEEEbiography}
[{\includegraphics[width=1in,height=1.25in,clip,keepaspectratio]{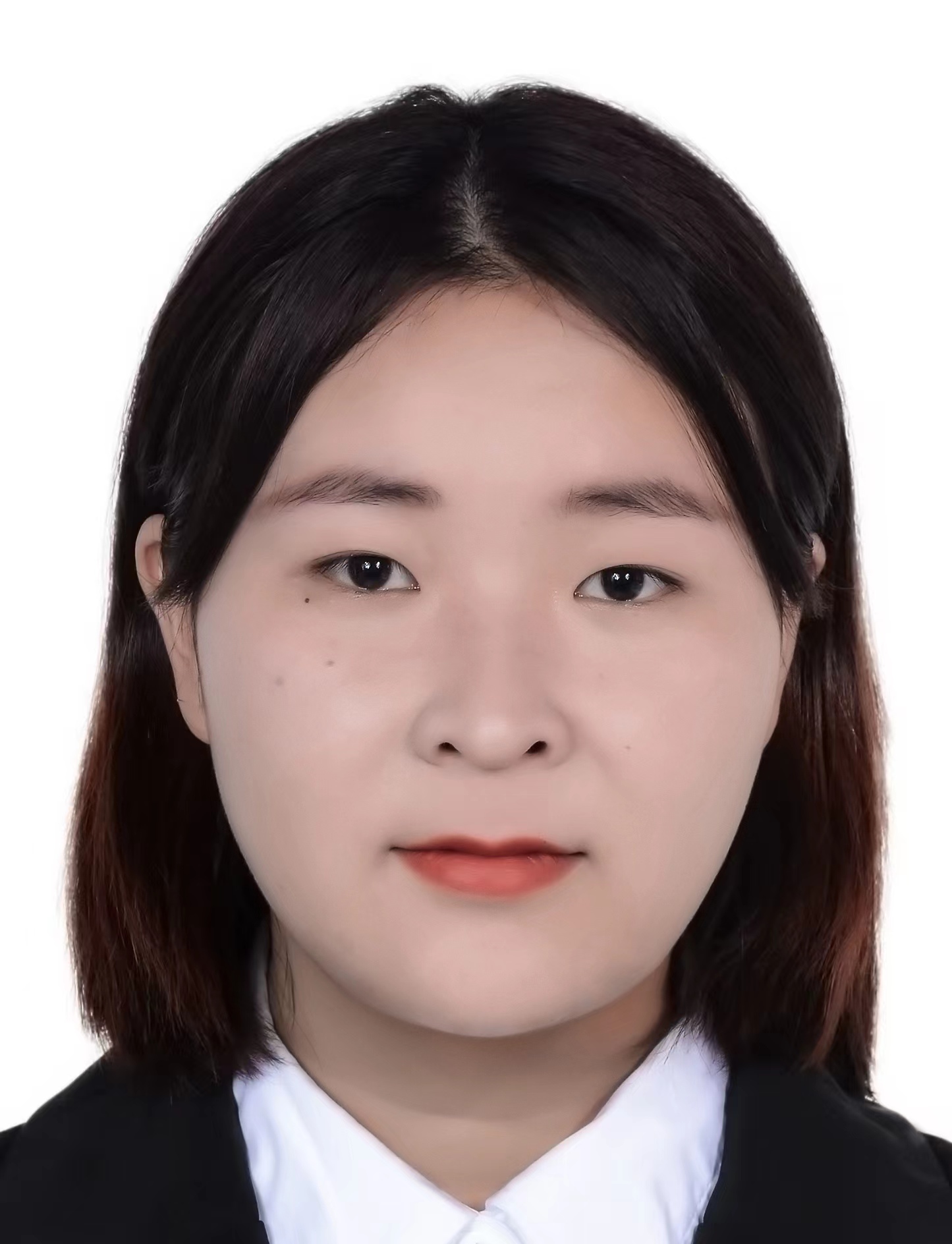}}]{Jiucui Lu}received the B.S. degree in computer science and technology from Qingdao University, Qingdao, China, in 2021. She is currently pursuing the M.S. degree with the Department of Computer Science and Technology, Ocean University of China, Qingdao, China. Her research
interests include digital image forensics and adversarial attacks.
\end{IEEEbiography}

\begin{IEEEbiography}[{\includegraphics[width=1in,height=1.25in,clip,keepaspectratio]{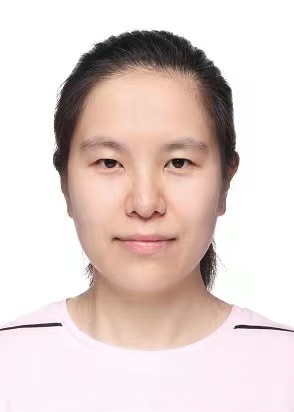}}]{Jiaran Zhou}
received the B.Eng. and Ph.D. degrees in Computer Science and Technology from Shandong University in 2012 and 2020, respectively. She is currently a Lecturer with the Center on Artificial Intelligence, Ocean University of China. Her research interests include computer graphics, geometry processing and artificial Intelligence security.
\end{IEEEbiography}

\vspace{-0.2cm}

\begin{IEEEbiography}[{\includegraphics[width=1in,height=1.25in,clip,keepaspectratio]{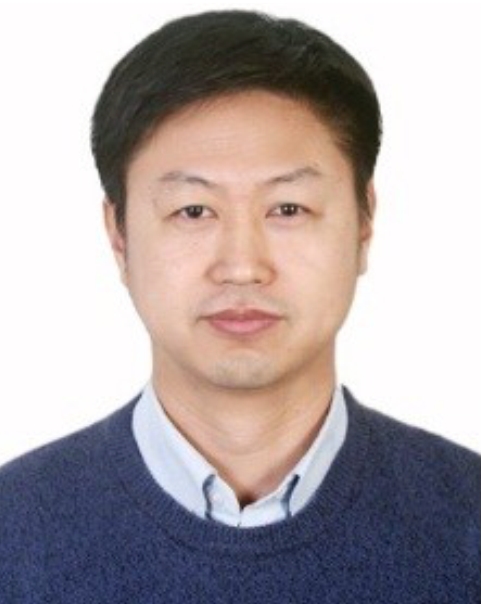}}]{Junyu Dong} (Member, IEEE) received the B.Sc.
and M.Sc. degrees in applied mathematics from the
Department of Applied Mathematics, Ocean University of China, Qingdao, China, in 1993 and 1999, respectively, and the Ph.D. degree in image processing from the Department of Computer Science, Heriot-Watt University, Edinburgh, U.K., in November 2003. He is currently a Professor and the Head of the Department of Computer Science and Technology. His research interests include machine learning, big data, computer vision, and underwater image processing.
\end{IEEEbiography}

\vspace{-0.2cm}

\begin{IEEEbiography}[{\includegraphics[width=1in,height=1.25in,clip,keepaspectratio]{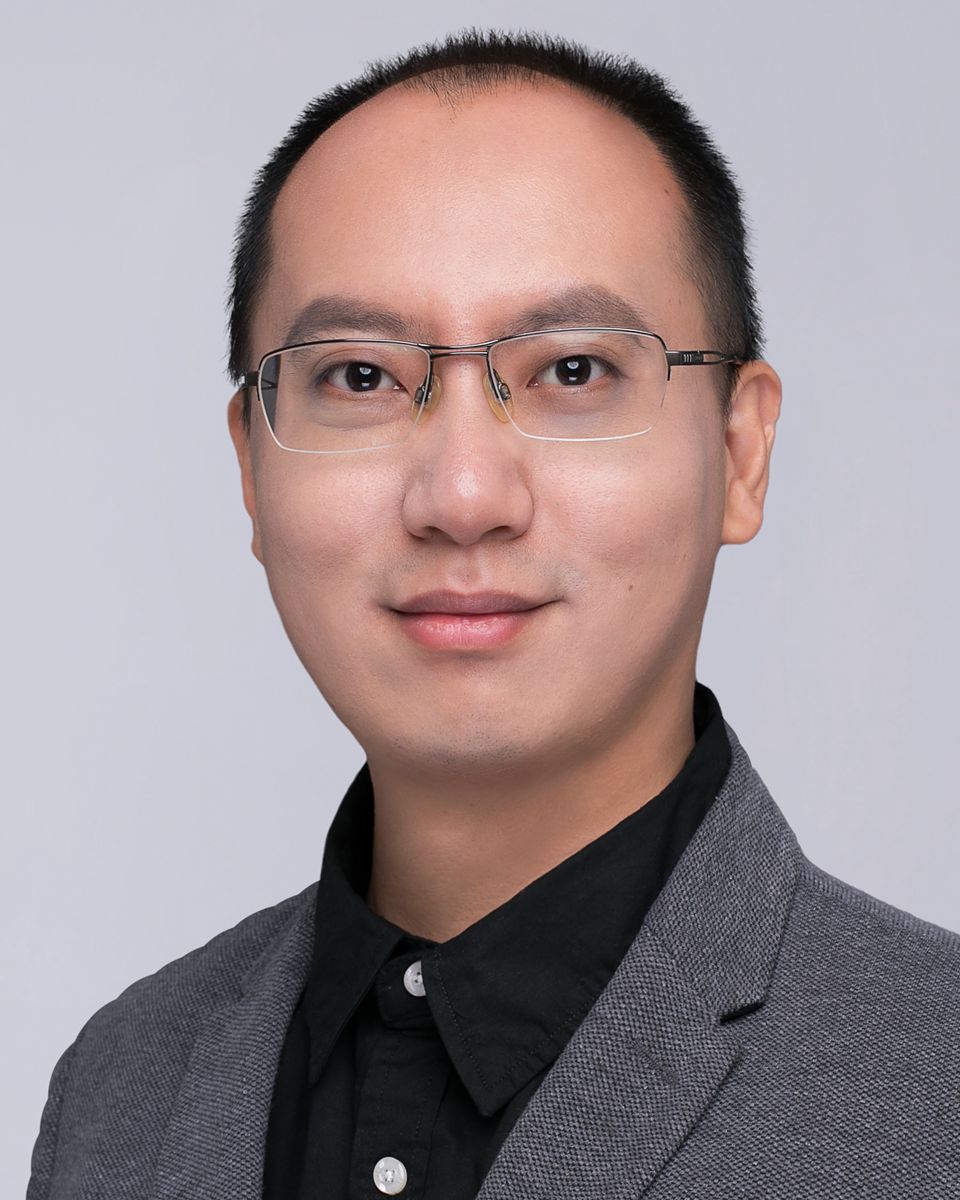}}]{Bin Li}
(Senior Member, IEEE) received the B.E. degree in communication engineering and the Ph.D. degree in communication and information system from Sun Yat-Sen University, Guangzhou, China, in 2004 and 2009, respectively. He was a Visiting Scholar with the New Jersey Institute of Technology, Newark, NJ, USA, from 2007 to 2008. In 2009, he joined Shenzhen University, Shenzhen, China, where he is currently a Professor. He is also the Director of the Guangdong Key Laboratory of Intelligent Information Processing and the Shenzhen Key Laboratory of Media Security. His current research interests include multimedia forensics, image processing, and deep machine learning. He is a Senior Area Editor of the IEEE TRANSACTIONS ON INFORMATION FORENSICS AND SECURITY.
\end{IEEEbiography}

\vspace{-0.2cm}

\begin{IEEEbiography}[{\includegraphics[width=1in,height=1.25in,clip,keepaspectratio]{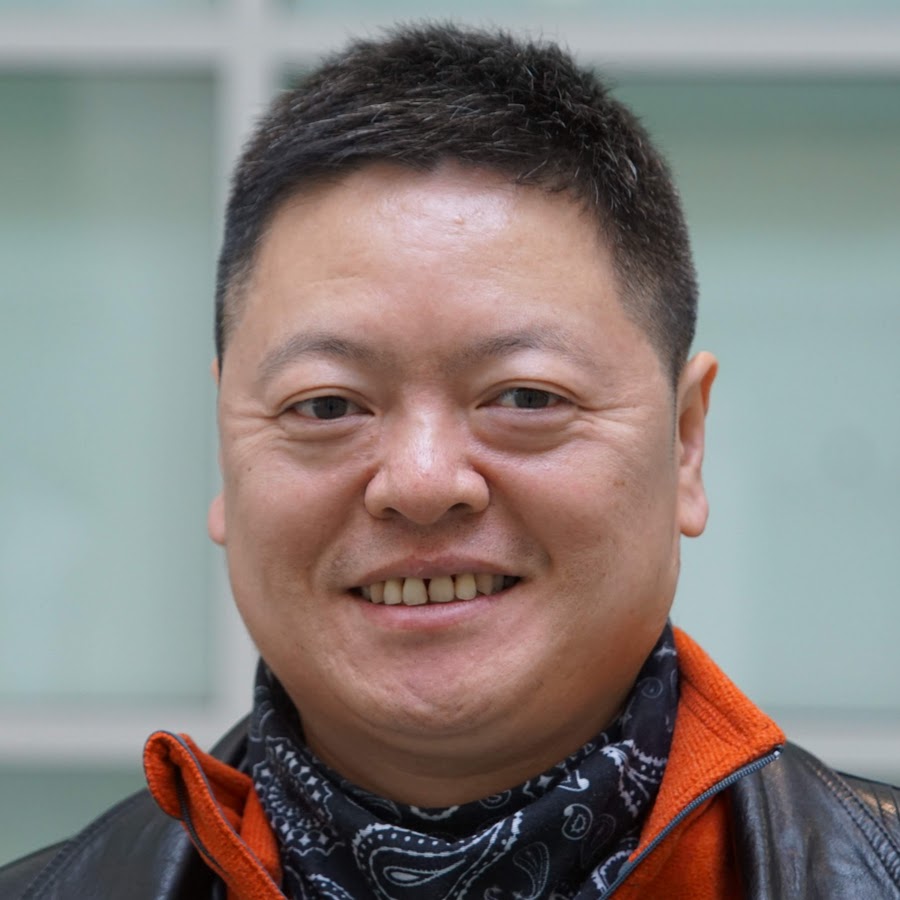}}]{Siwei Lyu}
(Fellow, IEEE) is a SUNY Empire Innovation Professor at the Department of Computer Science and Engineering, the University at Buffalo, State University of New York, USA. He earned his Ph.D. in Computer Science from Dartmouth College in 2005 and both his M.S. (2000) and B.S. (1997) degrees in Computer Science and Information Science, respectively, from Peking University, China. Dr. Lyu's research interests include digital media forensics, computer vision, and machine learning. He has published over 200 refereed journal and conference papers. 
\end{IEEEbiography}

\vspace{-0.2cm}

\begin{IEEEbiography}[{\includegraphics[width=1in,height=1.25in,clip,keepaspectratio]{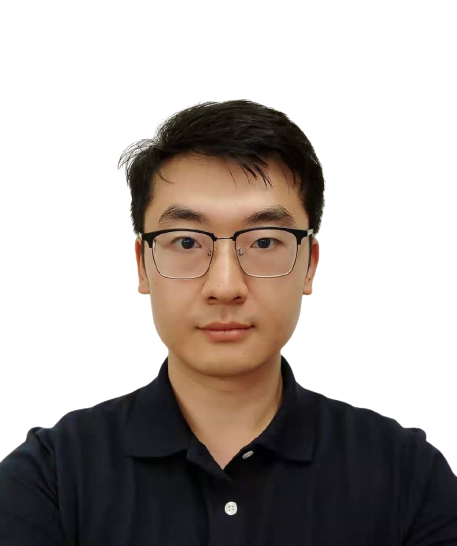}}]{Yuezun Li}
(Member, IEEE) received the B.S. degree in software
engineering from Shandong University in 2012, the
M.S. degree in computer science in 2015, and the
Ph.D. degree in computer science from the University at Albany, SUNY, in 2020. He was a Senior Research Scientist with the Department of Computer Science and Engineering, University at Buffalo, SUNY. He is currently a Lecturer with the Center on Artificial Intelligence, Ocean University of China. His work has been published in peer-reviewed conferences and journals, including ICCV, CVPR, TCSVT, PR, etc. His research interests include computer vision and multimedia forensics.
\end{IEEEbiography}

\end{document}